\documentclass{article}

\PassOptionsToPackage{numbers}{natbib}

\usepackage[preprint,nonatbib]{neurips_2023}

\usepackage[utf8]{inputenc} 
\usepackage[T1]{fontenc}    
\usepackage{hyperref}       
\usepackage{url}            
\usepackage{booktabs}       
\usepackage{amsfonts}       
\usepackage{nicefrac}       
\usepackage{microtype}      
\usepackage{xcolor}         
\usepackage{algorithmic}
\usepackage{graphicx}
\usepackage{amsmath,amssymb,amsfonts,amsthm}
\usepackage{microtype}

\usepackage[natbib,hyperref,sorting=none]{biblatex}
\title{DAGrid: Directed Accumulator Grid}

\author{%
  Hang Zhang \\
  Cornell University \\
  \texttt{hz459@cornell.edu} \\
  \And
  Renjiu Hu \\
  Cornell University \\
  \texttt{rh656@cornell.edu} \\
  \And
  Xiang Chen \\
  University of Leeds\\
  \texttt{X.Chen2@leeds.ac.uk} \\
  \And
  Rongguang Wang \\
  University of Pennsylvania \\
  \texttt{rgw@seas.upenn.edu} \\
  \And
  Jinwei Zhang \\
  Cornell University \\
  \texttt{jz853@cornell.edu} \\
  \And
  Jiahao Li \\
  Cornell University \\
  \texttt{jl3838@cornell.edu} \\
}

\begin{document}

\maketitle

\begin{abstract}

Recent research highlights that the Directed Accumulator (DA), through its parametrization of geometric priors into neural networks, has notably improved the performance of medical image recognition, particularly in situations confronted with the challenges of small and imbalanced datasets.
Despite the impressive results of DA in certain applications, its potential in tasks requiring pixel-wise dense predictions remains largely unexplored. 
To bridge this gap, we present the Directed Accumulator Grid (DAGrid), an innovative approach allowing geometric-preserving filtering in neural networks, thus broadening the scope of DA's applications to include pixel-level dense prediction tasks.
DAGrid utilizes homogeneous data types in conjunction with designed sampling grids to construct geometrically transformed representations, retaining intricate geometric information and promoting long-range information propagation within the neural networks. 
Contrary to its symmetric counterpart, grid sampling, which might lose information in the sampling process, DAGrid aggregates all pixels, ensuring a comprehensive representation in the transformed space.
The parallelization of DAGrid on modern GPUs is facilitated using CUDA programming, and also back propagation is enabled for deep neural network training. 
Empirical results clearly demonstrate that neural networks incorporating DAGrid outperform leading methods in both supervised skin lesion segmentation and unsupervised cardiac image registration tasks.
Specifically, the network incorporating DAGrid has realized a 70.8\% reduction in network parameter size and a 96.8\% decrease in FLOPs, while concurrently improving the Dice score for skin lesion segmentation by 1.0\% compared to state-of-the-art transformers.
Furthermore, it has achieved improvements of 4.4\% and 8.2\% in the average Dice score and Dice score of the left ventricular mass, respectively, indicating an increase in registration accuracy for cardiac images.
These advancements in performance indicate the potential of DAGrid for further exploration and application in the field of medical image analysis.
The source code is available at \url{https://github.com/tinymilky/DeDA}.

\end{abstract}
\section{Introduction}

Despite the successful application of neural networks in diverse medical image tasks such as physics-based inverse problems \cite{zhang2020fidelity,aggarwal2018modl}, deformable medical image registration \cite{balakrishnan2019voxelmorph,chen2021deep}, and lesion segmentation \cite{zhang2021all,wu2022fat}, adapting a well-established backbone architecture \cite{simonyan2014very,he2016deep,liu2021swin,dosovitskiy2020image,kirillov2023segment} to different tasks is often challenging.
This challenge stems not only from the domain shift \cite{wang2022metateacher,matsoukas2022makes} due to task variations and diverse acquisition protocols, but more significantly from the domain-specific nature and data limitations.
These conditions often lead to a failure of standard networks to extract unique task-specific information, such as the geometric structure of the target object when data is scarce.
Therefore, considering the domain-specific nature of medical images associated with different diseases, the challenge of how to incorporate useful inductive biases (priors) into neural networks for enhanced medical image processing remains unresolved.

Many medical imaging tasks often involve processing a primary target object, such as the left ventricle in cardiac image registration \cite{chen2021deep} or white matter hyperintensities in brain lesion segmentation \cite{zhang2023spatially}. 
These objects typically present strong geometric patterns, necessitating specialized image transformation techniques to capture these patterns. 
Directly incorporating geometric priors into the network can help mitigate the limitations of plain neural networks which often struggle to capture such patterns without ample training data.
The spatial transformer \cite{jaderberg2015spatial}, also known as Differentiable Grid Sampling (GS), is a learnable network module that facilitates image transformation within the neural network. 
Given a source feature map $\mathbf{U} \in \mathbb{R}^{C\times H \times W}$, a sampling grid $\mathbf{G} \in \mathbb{R}^{2\times H' \times W'}=(\mathbf{G}^x, \mathbf{G}^y)$ that specifies pixel locations to extract from $\mathbf{U}$, and a kernel function $\mathcal{K}()$ that defines the image interpolation, the output value at a specific position $(i,j)$ in the target feature map $\mathbf{V} \in \mathbb{R}^{C\times H' \times W'}$ can be expressed as: $\mathbf{V}_{ij}^{c} = \sum_n^{H}\sum_m^{W} \mathbf{U}_{nm}^{c}\mathcal{K}(\mathbf{G}_{ij}^x,n)\mathcal{K}(\mathbf{G}_{ij}^y,m)$.
The process described by the equation can be denoted as a function of tensor mapping, $\mathcal{S}(\mathbf{U};\mathbf{G},\mathcal{K}): \mathbb{R}^{C\times H \times W} \rightarrow \mathbb{R}^{C \times H' \times W'}$, where $\mathbf{U},\mathbf{G}$, and $\mathcal{K}$ represent the source feature map, sampling grid, and sampling kernel, respectively. 
Here, $H\times W$ denotes the spatial size of $\mathbf{U}$, while $H'\times W'$ represents the spatial size of $\mathbf{V}$.

\begin{figure}[!t]
	\centering
	\includegraphics[width=0.75\columnwidth]{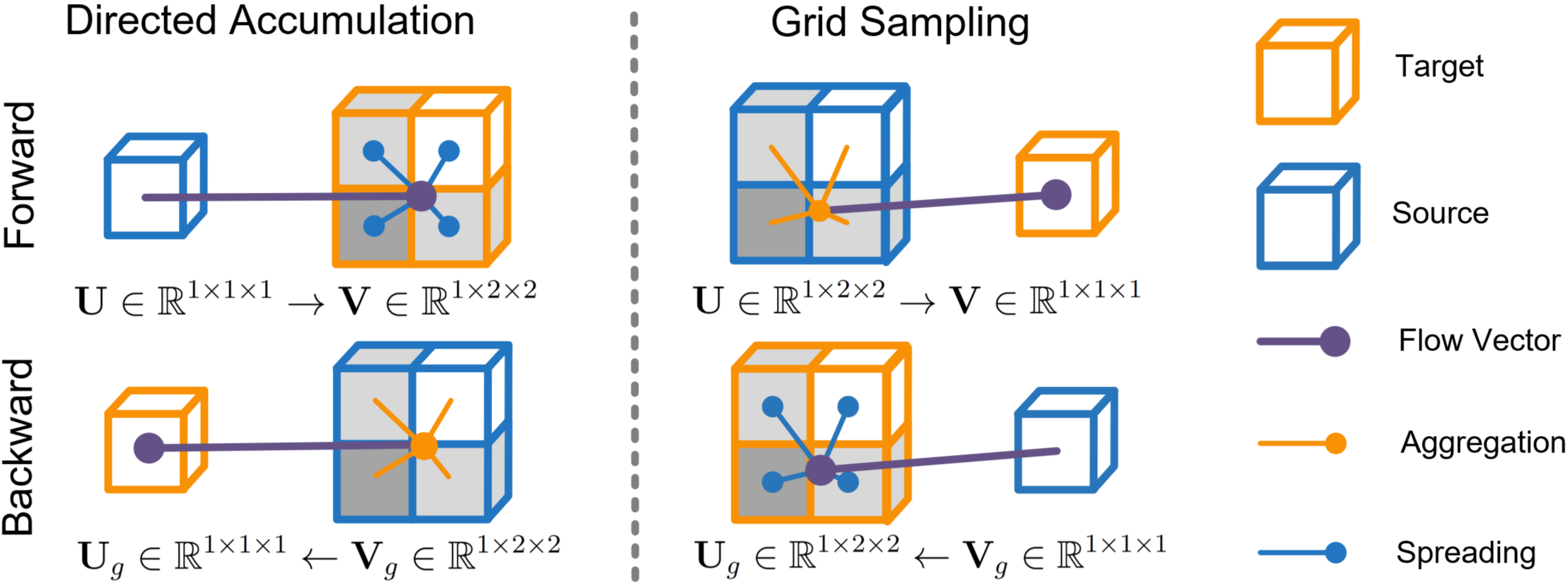}
	\caption{
        The above provides a visual illustration of the difference between DA and GS using bilinear sampling kernel.
        In the figure, $\mathbf{U}$ and $\mathbf{V}$ represent the input and output feature maps, respectively, while $\mathbf{U}_g$ and $\mathbf{V}_g$ denote the gradient maps of the input and output. 
        Notably, the figure highlights that the structure of data flow in the backward pass of DA mirrors that of the forward pass of GS. 
        This demonstrates that within the forward-backward learning framework, GS operates as a symmetrical counterpart to DA.
        } 
	\label{fig:da-gs}
        \vspace{-3ex}
\end{figure}

While GS has been successfully employed across a variety of vision applications, such as deformable medical image registration \cite{chen2021deep,balakrishnan2019voxelmorph}, object detection \cite{han2021redet}, optical flow estimation \cite{teed2020raft}, image classification \cite{esteves2018polar}, and image translation \cite{kong2021breaking}, it falls short in handling a specific class of image transformations that involve Directed Accumulation (DA) \cite{zhang2023deda}.
Contrary to GS, where the transformed representation may risk information loss during the sampling process when the mapping is not one-to-one, DA forms a new representation by aggregating information from all pixels within the feature map.
With a slight adaptation of notations, we can formulate DA as: $\mathbf{V}_{ij}^{c} = \sum_n^{H}\sum_m^{W} \mathbf{U}_{nm}^{c}\mathcal{K}(\mathbf{G}_{nm}^x,i)\mathcal{K}(\mathbf{G}_{nm}^y,j)$.
It's worth highlighting the difference between GS and DA: In GS, the sampling grid $\mathbf{G}$ and the target feature map $\mathbf{V}$ share the same spatial dimensions and coordinates are iterated over in the loop. 
Conversely, in DA, $\mathbf{G}$ shares the same spatial dimensions as the source feature map $\mathbf{U}$ and it's the elements of the sampling grid that are iterated over in the loop, as illustrated in Fig.~\ref{fig:da-gs}.
Given these modifications, DA allows for the parameterization of geometric shapes like rims \cite{zhang2022qsmrim} into neural networks as learnable modules.

In this paper, we present the Directed Accumulator Grid (DAGrid), which has been specifically designed to extend the applications of DA to pixel-level dense prediction tasks.
DAGrid consists of three main components: grid creation, grid processing, and grid slicing (terminology that aligns with \cite{paris2006fast}).
Through the coordinated functioning of these three components, DAGrid endows neural networks with three valuable characteristics for modeling geometric shapes: it enables explicit long-range information propagation, retains more information from the original image, and facilitates geometric-preserving filtering. 
These properties significantly enhance the capability of neural networks in handling medical tasks involving geometric patterns.
We have implemented DAGrid on modern GPUs using CUDA programming and have enabled back propagation for deep neural network training.


Here we highlight the effectiveness of DAGrid across two dense image mapping medical applications: skin lesion segmentation and cardiac image registration. 
Our contributions are threefold.
Firstly, we present DAGrid, an extension of DA that incorporates geometric filtering within neural networks, thereby expanding its utility to dense pixel-wise mapping tasks.
Secondly, we instantiate DAGrid as a circular accumulator (DA-CA) for cardiac image registration, demonstrating its ability to enhance the registration of inner-objects with specific geometric patterns. 
This resulted in a marked improvement of 4.4\% in the average Dice score and 8.2\% in the Dice score of the left ventricular mass.
We also employ DAGrid as a polar accumulator (DA-PA) in skin lesion segmentation tasks. 
Compared to conventional methods, DA-PA excels in preserving detailed information and outperforms transformer-based networks, achieving a 1.0\% improvement in Dice score, along with a substantial reduction of 70.8\% in network parameter size and 96.8\% in FLOPs.

\section{Related Work}

{\bf Learning to Accumulate} 
The Hough Transform (HT) \cite{osti4746348}, along with its subsequent variants or improvements, are widely used methods that leverage the value accumulation process and have been further enhanced in the neural network framework. 
Deep Voting \cite{xie2015deep} employs neural networks to generate Hough votes for nucleus localization in microscopy images. 
Hough-CNN \cite{milletari2017hough} applies Hough voting to improve MRI and ultrasound image segmentation performance. 
Network-predicted Hough votes \cite{qi2019deep,qi2020imvotenet} have achieved state-of-the-art performance in object detection within 3D point clouds. 
Memory U-Net \cite{zhang2021memory} utilizes CNNs to generate Hough votes for lesion instance segmentation. 
The central idea behind these methods is to use learning models to produce Hough votes, which maps local evidence to an application-specific transformed space.

{\bf Learning in Accumulation}
Local Convolution filters in the transformed space, also known as accumulator space, aggregates structural features such as lines \cite{lin2020deep,zhao2021deep} and rims \cite{zhang2023deda} in the image space, facilitating the incorporation of priors into networks. 
This accumulator space convolution, as opposed to attention-based methods \cite{wang2018non,zhang2021efficient}, explicitly captures long-range information through direct geometric parameterization.
Examples of this approach include Lin \emph{et al.} \cite{lin2020deep}, who use line parameterization as a global prior for straight line segmentation, and Zhao \emph{et al.} \cite{zhao2021deep}, who integrate the accumulator space into the loss function for enhanced semantic line detection.
Interestingly, semantic correspondence detection has seen improvements in both 3D point clouds \cite{lee2021deep} and 2D images \cite{min2019hyperpixel} through the use of convolutions in the accumulator space. 
Zhao \emph{et al.} \cite{zhao20223d} utilize HT to combine the Manhattan world assumption and latent features for 3D room layout estimation.
Originally developed by Chen \emph{et al.} \cite{chen2007real} to accelerate the bilateral filter \cite{tomasi1998bilateral}, the bilateral grid has been further employed in neural networks for applications such as scene-dependent image manipulation \cite{gharbi2017deep} and stereo matching \cite{xu2021bilateral}.

{\bf Learning with Geometric Priors}
The requirement of substantial datasets for training deep networks \cite{deng2009imagenet,lin2014microsoft,kirillov2023segment} poses challenges, especially for certain data-limited clinical applications. 
For instance, despite being trained on some of the largest datasets available, large vision model SAM \cite{kirillov2023segment} still underperform specialized models in many medical imaging tasks, even with fine-tuning \cite{ma2023segment}.
In contrast, incorporating geometric or domain-specific priors can be advantageous. 
Techniques such as distance transformation mapping \cite{ma2020distance} and spatial information encoding \cite{liu2018intriguing} have been successfully used to develop edge-aware loss functions \cite{kervadec2019boundary,9434085,karimi2019reducing}, network layers with anatomical coordinates \cite{zhang2021all} as priors, and spatially covariant network weight generation \cite{zhang2023spatially}.
Polar or log polar features have found wide application in tasks such as modulation classification \cite{teng2020accumulated}, rotation- and scale-invariant polar transformer networks \cite{esteves2018polar}, object detection \cite{xie2020polarmask,xu2019explicit,park2022eigencontours}, correspondence matching \cite{ebel2019beyond}, and cell detection \cite{schmidt2018cell} and segmentation \cite{stringer2021cellpose}.
Moreover, explicit geometric shapes like straight lines, concentric circles, and rims have facilitated semantic line detection \cite{lin2020deep,zhao2021deep}, rim lesion identification \cite{zhang2023deda}, and lithography hotspot detection \cite{zhang2016enabling,zhang2017bilinear}.
\section{Methods}

In this section, we delineate the formulation of DAGrid. 
This differentiable module initiates by transforming an input feature map based on the specific sampling grids into a transformed accumulator space, referred to as grid space. 
Subsequently, operations such as convolution are performed in this grid space, followed by slicing back to the original feature map. 
In the case of multi-channel input, the same transformation process is applied to each channel. 
The DAGrid is composed of three components: grid creation, grid processing, and grid slicing. 
The synergy of these components facilitates geometric-preserving filtering, thus enhancing medical applications that rely on geometric priors.

\subsection{DAGrid Creation}

Given a source feature map $\mathbf{U} \in \mathbb{R}^{C\times H \times W}$, a target feature map $\mathbf{V} \in \mathbb{R}^{C\times H' \times W'}$, a set of sampling grids $\mathcal{G} = \{\mathbf{G}[k] \in \mathbb{R}^{2\times H \times W}=(\mathbf{G}^x[k], \mathbf{G}^y[k])~|~k \in \mathbb{Z}^+, 1 \leq k \leq N \}$ ($N\geq 1$ is the number of grids), and a kernel function $\mathcal{K}()$, the output value of a particular cell $(i,j)$ at the target feature map $\mathbf{V}$ can be written as follows:
\begin{equation}
    \mathbf{V}_{ij}^{c} = \sum_k^{N}\sum_n^{H}\sum_m^{W} \mathbf{U}_{nm}^{c}\mathcal{K}(\mathbf{G}_{nm}^x[k],i)\mathcal{K}(\mathbf{G}_{nm}^y[k],j),
    \label{eq:deda}
\end{equation}
where the kernel function $\mathcal{K}()$ can be replaced with any other specified kernels, e.g. integer sampling kernel $\delta(\lfloor\mathbf{G}_{nm}^x+0.5\rfloor-i)\cdot \delta(\lfloor\mathbf{G}_{nm}^y+0.5\rfloor-j)$ and bilinear sampling kernel $\text{max}(0,1-|\mathbf{G}_{nm}^x-i|) \cdot \text{max}(0,1-|\mathbf{G}_{nm}^y-j|)$. 
Here $\lfloor x+0.5\rfloor$ rounds $x$ to the nearest integer and $\delta()$ is the Kronecker delta function.
The Eq.~\ref{eq:deda} can be denoted as a function mapping, $\mathcal{D}(\mathbf{U};\mathcal{G},\mathcal{K}): \mathbb{R}^{C\times H \times W} \rightarrow \mathbb{R}^{C \times H' \times W'}$.

\begin{figure}[!t]
	\centering
	\includegraphics[width=1\columnwidth]{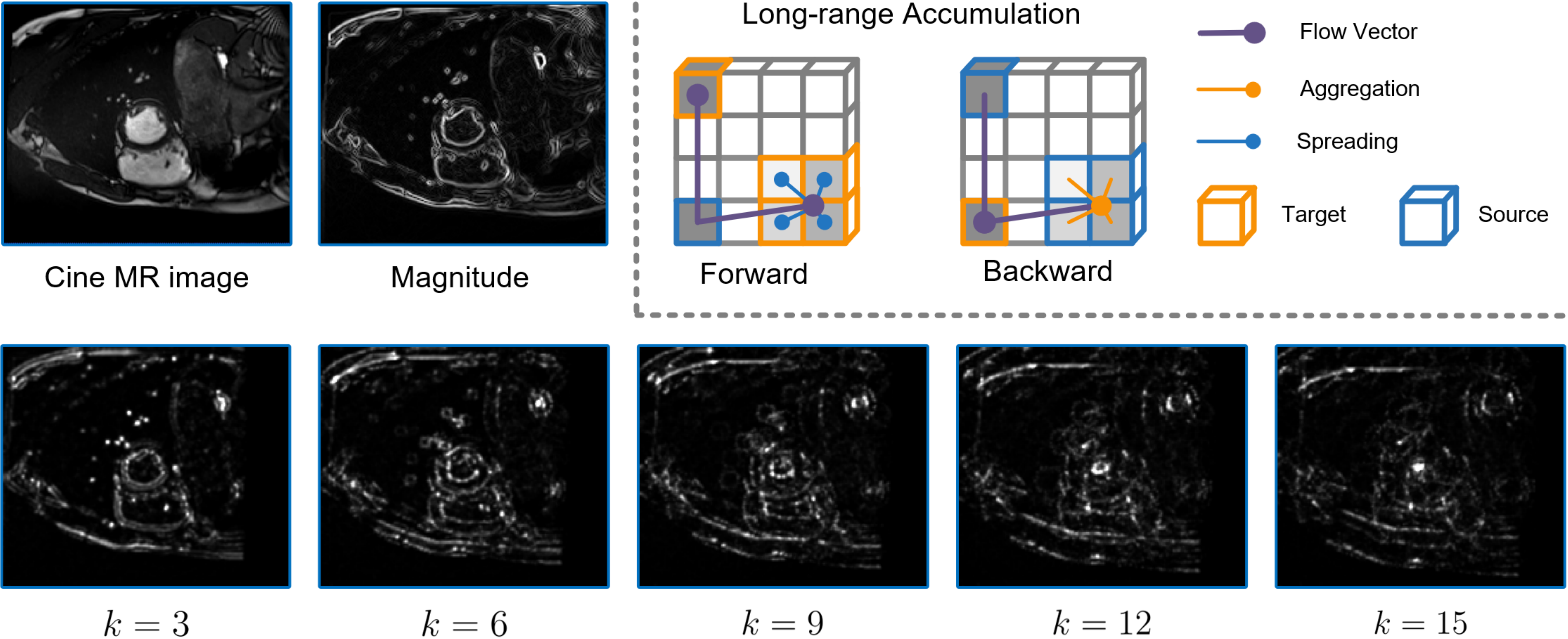}
	\caption{
        The above provides a visual illustration of the explicit long-range information propagation in DAGrid. 
        The top right panel presents a toy example that demonstrates how values and gradients propagate between the forward and backward pass during network training. 
        The remaining figures show the information propagation in a real cine MR image. To clarify the propagation process, we utilize the magnitude of the MR image and apply Eq.\eqref{eq:deda} with the sampling grids generated in Eq.\eqref{eq:hct_grids}. 
        It should be noted that we have limited $k$ to only one value for the purpose of visualization.
        } 
        \vspace{-2ex}
	\label{fig:long-range}
\end{figure}

To enable geometric-preserving filtering in the DAGrid, it's important to monitor the number of pixels (or a weight) corresponding to each grid cell. 
Thus, during grid creation, we store homogeneous quantities $(\mathbf{V}{ij}^{c}\cdot\mathbf{W}{ij}^{c},\mathbf{W}_{ij}^{c})$. 
Here, $\mathbf{W}$ can be derived from $\mathbf{W}=\mathcal{D}(\mathbf{J};\mathcal{G},\mathcal{K})$, where $\mathbf{J}$ is a tensor of ones. 
This representation simplifies the computation of weighted averages: $(w_1v_1,w_1)+(w_2v_2,w_2)=(w_1v_1+w_2v_2,w_1+w_2)$. 
Normalizing by the homogeneous coordinates $(w1+w2)$ yields the anticipated averaging of $v_1$ and $v_2$, weighted by $w_1$ and $w_2$.
Conceptually, the homogeneous coordinate $\mathbf{W}$ represents the importance of its associated data $\mathbf{V}$.


\subsection{DAGrid Processing and Slicing}

Any function $f$ that inputs a tensor and outputs another can process the accumulator grid $\tilde{\mathbf{V}}=f(\mathbf{V})$. 
There's no requirement for $\tilde{\mathbf{V}}$ and $\mathbf{V}$ to be of the same size, as long as it suits the grid slicing. 
For image processing, $f$ could be a bilateral filter, a Gaussian filter, or a non-maximal suppression operator. 
Within a neural network, $f$ could be a learnable convolutional layer or even a complete backbone network such as U-Net \cite{ronneberger2015u}.

After grid processing, we need to extract the feature map back by slicing.
Slicing is the critical DAGrid operation that yields piece-wise smooth output in terms of the geometric shape.
Given a processed accumulator grid $\tilde{\mathbf{V}}\in \mathbb{R}^{C\times H' \times W'}$ and a sampling grid set  $\mathcal{G} = \{\mathbf{G}[k] \in \mathbb{R}^{2\times H \times W}=(\mathbf{G}^x[k], \mathbf{G}^y[k])~|~k \in \mathbb{Z}^+, 1 \leq k \leq N \}$ (it is not necessary, but usually the set is the same as the set in DAGrid creation), the slicing can be formulated as follows:
\begin{equation}
    \tilde{\mathbf{U}}_{ij}^{c} = \sum_k^{N}\sum_n^{H'}\sum_m^{W'} \tilde{\mathbf{V}}_{nm}^{c}\mathcal{K}(\mathbf{G}_{ij}^x[k],n)\mathcal{K}(\mathbf{G}_{ij}^y[k],m),
    \label{eq:slice}
\end{equation}
where $\tilde{\mathbf{U}}$ is the feature map that has been sliced back, and we use notation $\mathcal{S}((\tilde{\mathbf{V}};\mathcal{G},\mathcal{K}): \mathbb{R}^{C\times H' \times W'} \rightarrow \mathbb{R}^{C \times H \times W})$ to signify Eq.~\ref{eq:slice}.
Regardless of the grid processing that occurs in the intermediate stages, it is evident that the processes represented by Eq.~\ref{eq:deda} and Eq.~\ref{eq:slice} are symmetrical to each other. 
It's the processing within the grid, between creation and slicing, that makes geometric-preserving operations feasible.

\begin{figure}[!t]
	\centering
	\includegraphics[width=1\columnwidth]{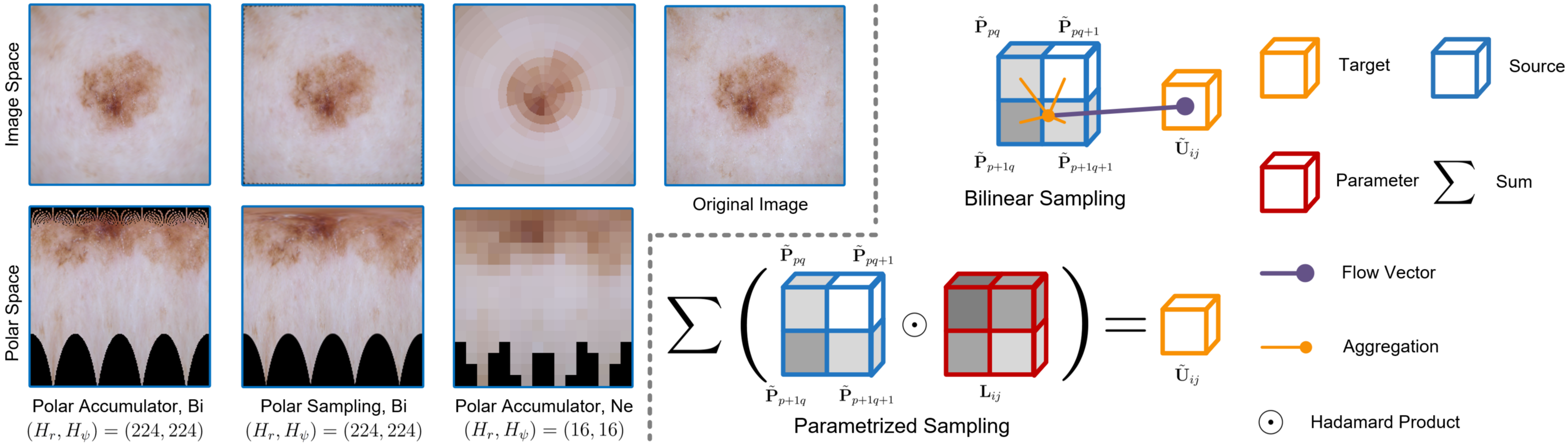}
	\caption{
        The above figure visually illustrates the creation of a polar accumulator grid. 
        The left panel depicts the polar accumulator and polar sampling with different parameters applied to a skin lesion image. 
        The right panel provides a detailed comparison between bilinear sampling and parametrized sampling. 
        Here, $\mathbf{L}_{ij} \in \mathbb{R}^{2\times 2}$ represents the parameter matrix associated with the output $\tilde{\mathbf{U}}_{ij}$, while $\tilde{\mathbf{P}}_{pq}$, $\tilde{\mathbf{P}}_{p+1q}$, $\tilde{\mathbf{P}}_{pq+1}$, and $\tilde{\mathbf{P}}_{p+1q+1}$ denote the four cells that originally were involved in the bilinear sampling process.
        } 
        \vspace{-1ex}
	\label{fig:param}
\end{figure}

\subsubsection{Circular and Polar Accumulator Grid}

The distinction between the circular, polar or any other accumulators originates from their respective sampling grids, which are essential in delineating the geometric transformation pertinent to specific applications.
In essence, we can establish a novel accumulator grid by defining a new geometric transformation via a set of custom sampling grids. 
Utilizing a bilinear sampling kernel, this set of sampling grids can also be learned or fine-tuned through back propagation. 
For simplicity, in the remainder of this paper, we will represent the feature map using spatial dimensions only.

{\bf Circular Accumulator Grid}
Let $\mathbf{U} \in \mathbb{R}^{H \times W}$ denote the input feature map. The magnitude of image gradients can be calculated as $\mathbf{S} = \sqrt{\mathbf{U}_x\odot\mathbf{U}_x + \mathbf{U}_y\odot\mathbf{U}_y}$, where $\odot$ represents the Hadamard product, $\mathbf{U}_x = \dfrac{\partial \mathbf{U}}{\partial x}$, and $\mathbf{U}_y = \dfrac{\partial \mathbf{U}}{\partial y}$. Image gradient tensors $\mathbf{U}_x$ and $\mathbf{U}_y$ can be efficiently calculated using convolution kernels like the Sobel operator.
Normalized gradients can be obtained as $\hat{\mathbf{U}}_x = \dfrac{\mathbf{U}_x}{\mathbf{S}+\epsilon}$ and $\hat{\mathbf{U}}_y = \dfrac{\mathbf{U}_y}{\mathbf{S}+\epsilon}$, where $\epsilon$ is a small real number added to avoid division by zero. 
Mesh grids of $\mathbf{U}$ are denoted as $\mathbf{M}_x$ (value range: $(0,H-1)$) and $\mathbf{M}_y$ (value range: $(0,W-1)$).
A set of sampling grids can be generated as
\begin{equation}
    \mathcal{G} = \{\mathbf{G}[k]=(\mathbf{G}^x[k], \mathbf{G}^y[k])~|~k \in \mathbb{Z}^+, 1 \leq k \leq N \},
    \label{eq:hct_grids}
\end{equation}
where $\mathbf{G}[k] \in \mathbb{R}^{2\times H \times W}$, $\mathbf{G}^x[k]=k\hat{\mathbf{U}}_x+\mathbf{M}_x$, $\mathbf{G}^y[k]=k\hat{\mathbf{U}}_y+\mathbf{M}_y$, and $N=\text{max}(H,W)$.
Let $\mathcal{G}^{-}$ denote the set of sampling grids with gradients in the opposite direction, where $\mathbf{G}^x[k]^{-}=-k\hat{\mathbf{U}}_x+\mathbf{M}_x$, $\mathbf{G}^y[k]^{-}=-k\hat{\mathbf{U}}_y+\mathbf{M}_y$.
The circular accumulator grid can be formulated as $\mathbf{V}_s = \mathcal{D}(\mathbf{S};\mathcal{G},\mathcal{K})-\mathcal{D}(\mathbf{S};\mathcal{G}^{-},\mathcal{K})$ and $\mathbf{V}_u = \mathcal{D}(\mathbf{U};\mathcal{G},\mathcal{K})-\mathcal{D}(\mathbf{U};\mathcal{G}^{-},\mathcal{K})$, where $\mathbf{V}_s$ and $\mathbf{V}_u$ represent the accumulated feature and magnitude value, respectively.
In this scenario, the bilinear sampling kernel is used to track gradients for the sampling grids.

{\bf Polar Accumulator Grid}
Let $\mathbf{U} \in \mathbb{R}^{H \times W}$ be the input feature map, $\mathbf{M}^x \in \mathbb{R}^{H\times W}$ (value range: $(0,H-1)$) and $\mathbf{M}^y \in \mathbb{R}^{H\times W}$ (value range: $(0,W-1)$) be the corresponding mesh grids, $(x_c,y_c)$ be the coordinate of image center,
Let the mesh grids of $\mathbf{U}$ be $\mathbf{M}^x$ (value range: $(0,H-1)$) and $\mathbf{M}^y$ (value range: $(0,W-1)$), the coordinate of image center be $(x_c,y_c)$, the value of sampling grid in the radial direction $\mathbf{G}^{x}$ at position $(i,j)$ can be obtained as: $\mathbf{G}^{x}_{ij} = \sqrt{(\mathbf{M}^x_{ij} - x_c)^2 + (\mathbf{M}^y_{ij} - y_c)^2} / s_r$, where $s_r$ is the sampling rate in the radial direction.
Similarly, the value of sampling grid in the angular direction $\mathbf{G}^{y}$ at position $(i,j)$ can be obtained as: $\mathbf{G}^{y}_{ij} = \arctan{((\mathbf{M}^x_{ij} - x_c)^2 + (\mathbf{M}^y_{ij} - y_c)^2} + \pi) / s_{\theta}$, where $s_{\theta}$ is the sampling rate in the angular direction, and addition of $\pi$ is to shift all values into the range of $(0,2\pi)$.
The process of polar accumulation for each channel is the same, requiring just one sampling grid. 
Given $\mathcal{G}\{\mathbf{G}=(\mathbf{G}^{x},\mathbf{G}^{y})\}$, we can generate the polar accumulator grid through Eq.~\ref{eq:deda} as $\mathbf{P}=\mathcal{D}(\mathbf{U};\mathcal{G},\mathcal{K}) \in \mathbb{R}^{H_r\times W_\psi}$, where $H_r$ and $W_\psi$ are determined by the sampling rate $s_r$ and $s_\theta$.
After processing $\mathbf{P}$ with a neural network $f$, we can slice the processed grid $\tilde{\mathbf{P}}=f(\mathbf{P})$ back into the image space as $\tilde{\mathbf{U}}=\mathcal{S}(\tilde{\mathbf{P}};\mathcal{G},\mathcal{K})$.
It's important to note that before processing $\mathbf{P}$ with $f$, we use homogeneous coordinates for normalization.




\subsection{DAGrid Functionality}

Before delving into the experiments, we intuitively demonstrate three valuable characteristics of DAGrid for modeling geometric shapes in neural networks: explicit long-range information propagation, retaining more information from the original image, and geometric-preserving filtering.

{\bf Explicit Long-Range Information Propagation}
As illustrated in the top right panel of Fig.~\ref{fig:long-range}, during the forward pass, DAGrid is capable of transferring values from the input feature map to specific cells in the accumulator grid using the sampling grids. 
Subsequently, during the backward pass, gradient values initially found in the specified accumulator cell flow along the same route to the input feature map. 
This process allows DAGrid to achieve explicit long-range information propagation.
As illustrated in Fig.~\ref{fig:long-range}, with the increase in radius from 3 to 15, the magnitude information progressively propagates towards the center of the left ventricle.

{\bf Retaining more Information from the Original Image}
In traditional grid sampling, each cell in the output feature map derives its value from the corresponding cell in the input feature map. 
This might lead to potential information loss, particularly when the mapping between input and output is not one-to-one. 
Conversely, in the directed accumulation process, all pixel values from the input are used to construct the accumulator grid. 
While the values undergo smoothing during the normalization of homogeneous coordinates, this method ensures that all values from the input contribute to the accumulation process, providing the potential for information recovery. 
By harnessing the power of neural networks, we can augment the slicing process with implicit parametrization, substituting the conventional bilinear sampling with a learnable linear combination module.
This adjustment facilitates optimal utilization of the preserved information, leading to an improvement in the overall performance of the model.

Let $\tilde{\mathbf{P}}$ be the processed polar grid, $p=\lfloor \mathbf{G}{ij}^x \rfloor$, $q=\lfloor \mathbf{G}{ij}^y \rfloor$, we can parametrize the slicing process of polar grid with the incorporation of a learnable linear combination module as follows:
\begin{equation}
    \tilde{\mathbf{U}}_{ij} = \sum_{n=p}^{p+1}\sum_{m=q}^{q+1} \tilde{\mathbf{P}}_{nm}\mathbf{L}_{ij}[n-p][m-q],
    \label{eq:param}
\end{equation}
where $\mathbf{L} \in \mathbb{R}^{H\times W \times 2 \times2}$ is a parameter tensor that can be learned during the network training. 
With Eq.~\ref{eq:param}, the slicing process transcends the constraints of bilinear sampling, allowing the network to learn how to sample the most valuable information from each of the four cells originally involved in the bilinear sampling process.
Indeed, during the accumulation phase, each cell in the accumulator grid becomes a weighted sum of certain cells in the input. 
Then, during the slicing phase, the learnable linear combination module is employed to recover as much valuable information as possible. 
This method assists in preserving the finer details of the original image.
Please see the right panel in Fig.~\ref{fig:param} for a visual illustration.

{\bf Geometric-Preserving Filtering}

In image processing, many useful image components or manipulations are often piece-wise smooth rather than purely band-limited \cite{paris2006fast,chen2007real,chan2001active}. 
For instance, the image segmentation produced by a neural network should be smooth within each segment, and the deformation field of image registration should be piece-wise smooth within each region \cite{nie2021deformable,chen2022joint}. 
Therefore, these components or manipulations can be accurately approximated by specially designed low-frequency counterparts.
Additionally, if these low-frequency counterparts possess geometric structures, slicing from them yields piece-wise smooth output that preserves such structure. 
Consider the third column in Fig.~\ref{fig:param} as an example: the polar accumulator employs nearest sampling using a grid size of $(H_r,H_\psi)=(16,16)$. 
Slicing from the accumulator grid results in piece-wise smooth regions along the angular and radial directions.
Furthermore, applying convolutional filters in the geometric transformed grid space equates to conducting convolutions with respect to the geometric structure in the image space. 
This process facilitates the performance of geometric-preserving operations during the processing within the grid, between creation and slicing.
\section{Experiments}

\begin{table*}[!t]
\caption{
Quantitative comparisons of skin lesion segmentation results produced by DA-PA and other comparator networks. 
The best performing metrics are bolded.}
\label{tab:segmentation}
\begin{center}
\resizebox{1.\columnwidth}{!}{
\begin{tabular}{ lcccccc }
\hline
\hline
 Model &  Avg. Dice (\%) & Avg. Precision (\%) & Avg. Sensitivity (\%) & Avg. Jaccard Index (\%) & FLOPs Ratio & Param Size Ratio \\
\hline
resU-Net \cite{zhang2018road} & 80.39 & 90.04 & 79.61 & 70.92 & 53.32 & 4.28\\
Tiramisu \cite{zhang2019multiple} & 85.09 & 85.75 & 89.67 & 76.08 & 44.39 & 0.24\\
All-Net \cite{zhang2021all} & 86.16 & 88.54 & 88.71 & 78.04 & 1.49 & 1.00 \\
FTN \cite{he2022fully} & 76.98 & 79.15 & 82.91 & 67.38 &2.04 &0.70\\
FAT-Net \cite{wu2022fat} & 88.49  & 87.20 & \textbf{92.92} & 80.67 & 31.13 & 3.42\\
DA-PA & \textbf{89.41} & \textbf{90.11} & 92.08 & \textbf{82.56} & 1.00 & 1.00\\
\hline
\end{tabular}
}
\end{center}
\vspace{-2ex}
\end{table*}

\subsection{Skin Lesion Segmentation}

In the first experiment, we compare the performance of our DAGrid-based polar accumulator (DA-PA) with other neural network-based methods.
In the case of convolutional neural networks (CNNs), we utilize the Tiramisu network \cite{zhang2019multiple}, which comprises densely connected blocks, the All-Net \cite{zhang2021all} with a tailored U-Net as the backbone network for lesion segmentation, and the residual U-Net (resU-Net)\cite{zhang2018road}.
For transformer networks, we employ the FAT-Net \cite{wu2022fat} with its feature adaptive block, and a fully transformer network \cite{he2022fully} designed for simultaneous skin lesion segmentation and classification.



\subsubsection{Datasets and Implementation details}

To evaluate the performance of the skin lesion segmentation, we use the public ISIC 2018 dataset \cite{codella2019skin} to compare performance of DA-CA with other CNN or transformer based networks.
ISIC 2018 dataset contains 2594 images for training, 100 images for validation, and 1000 images for testing.
Images and segmentation masks are resized to $(224,224)$ and then translated to align their center of mass to the geometric center for both training and testing.
Random flipping, random affine transformations, and random motion are applied to augment the data.

In our implementation, we utilize the backbone network from All-Net \cite{zhang2021all} for skin lesion segmentation learning.
To integrate the polar accumulator, several convolution blocks are introduced prior to the image transformation by the polar accumulator. 
This transformation is followed by the backbone network.
The output from the backbone network is then sliced back to the image space, followed by a few convolution blocks that output the segmentation logits.
For the polar grid accumulation, we employ bilinear sampling, and for polar grid slicing, we use parametrized sampling. 
We set $H_r=64$ and $H_\psi=64$.
For additional information regarding the network architecture and training details, please refer to the supplementary materials.

We evaluate the performance of each method using the average Dice score, Precision, Sensitivity and Jaccard Index. 
The higher of these scores indicate better performance.
In addition, we use DA-CA as base model to compute ratios of FLOPs and parameter size.
All ratios are computed by comparing the target model to the base model.
Table \ref{tab:segmentation} shows the ratios computed using an input tensor size of $1\times 3 \times 224 \times 224$.

\subsubsection{Results}

Table \ref{tab:segmentation} illustrates that DA-PA has outperformed other methods in terms of both the averaged Dice score and the Jaccard Index. 
Notably, DA-PA improves upon All-Net, the best CNN, by 3.8\% in Dice score and 5.8\% in the Jaccard Index. This improved performance is achieved despite DA-PA using the same backbone network as All-Net. Additionally, DA-PA's configuration, which includes a few added convolution blocks before and after the backbone and grid sizes $H_r=64$ and $H_\psi=64$, results in nearly the same parameter size as All-Net, but with a 49\% reduction in FLOPs. 
This is due to DA-PA's smaller input size to the backbone network, further enhancing its efficiency.

Furthermore, DA-PA outperforms the top-performing transformer network, FAT-Net, by improving the Dice score by 1.0\% and the Jaccard Index by 2.3\%, while concurrently achieving substantial reductions in FLOPs (96.8\%) and network parameter size (70.8\%). 
The superior performance of DA-PA can be attributed to its effectiveness in capturing the inherent characteristics of skin lesions: they typically exhibit different textures from their surroundings and generally display a shape that radiates from the center towards the periphery. 
This allows the low frequency components in the polar grid to adequately approximate the lesion geometry with less consideration of the texture. 
Moreover, a smaller polar grid size provides a larger receptive field and broader segments along the angular and radial directions in the image space, resulting in enhanced accuracy and efficiency.
For qualitative results and ablation study, please refer to the supplementary materials.

\begin{figure}[!t]
	\centering
	\includegraphics[width=1\columnwidth]{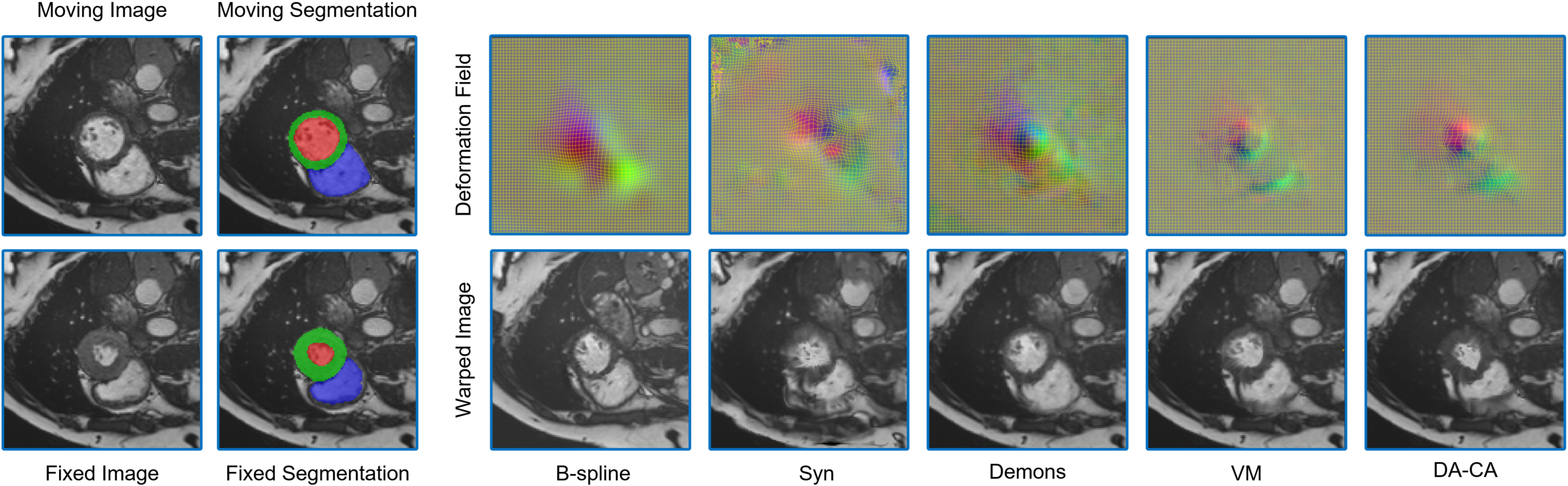}
	\caption{
        A qualitative comparison of cardiac image registration performance between our method and other comparative approaches.
        } 
	\label{fig:registration}
        \vspace{-2ex}
\end{figure}

\subsection{Cardiac Image Registration}

In the second experiment, we compare the performance of the DAGrid-based circular accumulator (DA-CA) with both traditional and deep learning-based registration methods. 
For traditional methods, we use B-splines registration (with a maximum of 1000 iteration steps and 1000 random points sampled per iteration), available in SimpleElastix~\cite{marstal2016simpleelastix}. 
We also apply Fast Symmetric Forces Demons \cite{vercauteren2009diffeomorphic} (with 100 iterations at standard deviations of 1.0) available in SimpleITK, and Symmetric Normalisation (SyN, using 3 resolution levels, with 100, 80, 60 iterations respectively) in ANTS~\cite{avants2011reproducible} as the baseline methods. 
In terms of deep learning-based approaches, we compare DA-CA with the VoxelMorph (VM) method~\cite{balakrishnan2019voxelmorph}.

\subsubsection{Datasets and Implementation details}
In this study, we utilize cine-MR images from the Automatic Cardiac Diagnosis Challenge (ACDC) dataset~\cite{bernard2018deep}. 
The ACDC dataset consists of 100 subjects, each with a complete cardiac cycle of cine MR images, along with corresponding segmentation masks available for end-diastole (ED) and end-systole (ES) frames. 
Our work focuses on intra-subject registration, specifically involving the left ventricle (LV), right ventricle (RV), and left ventricular mass (LVM).  
We register from the end-diastole (ED) frame to the end-systole (ES) frame or vice versa within the same subject, which gives us a total of 200 registration pairs or samples. 
These samples are subsequently split into training, validation, and testing sets, with 100, 50, and 50 samples respectively. 
All the cardiac MR images are re-sampled to a spacing of $1.8 \times 1.8 \times 10.0$, and then cropped to dimensions of $128 \times 128 \times 16$ pixels.

In our implementation, we adopt the backbone networks from VM \cite{balakrishnan2019voxelmorph} to learn the deformation field. 
However, our model has two branches: one branch is identical to VM, while the other introduces several convolution blocks before the image is transformed by the circular accumulator, which is then followed by the backbone. 
The output features from these two branches are then merged to compute the final deformation field. 
In the circular accumulator, we employ bilinear sampling and three radii ($N=15$, $N=10$, and $N=5$). 
More details regarding the network architecture can be found in the supplementary materials.

We evaluate the performance of each method using the average Dice score, Dice score for each separate region (i.e., LV, LVM, and RV), and the Hausdorff Distance (HD). 
A higher Dice score or a lower HD indicates better registration performance. 
Additionally, following \cite{chen2021deep}, we calculate two clinical cardiac indices, the LV end-diastolic volume (LVEDV) and LV myocardial mass (LVMM), to assess the consistency of cardiac anatomical structures after registration. 
These indices are computed based on the moving segmentation and warped moving segmentation. 
For clinical indices, values closer to the reference (computed based on the mixed and fixed segmentation) are considered better.

\subsubsection{Results}
{\bf Qualitative Results} 
The qualitative comparison between our method and other approaches is depicted in Fig.~\ref{fig:registration}.
It is evident that deep learning-based methods (VM and our DA-CA) outperform traditional methods (B-spline and Demons), in terms of the similarity between warped images and the fixed image.
It is worth noting that there is substantial distortion between the moving and fixed images due to cardiac contraction from ED to ES, which complicates the registration process.
While other methods struggle to accurately capture such distortion, DA-CA, with its ability for long-range information propagation, effectively addresses the changes around the LV. 
This results in a warped moving image that is most similar to the fixed image among all the methods tested.
In addition, from the deformation field, a more realistic motion pattern across all three regions of LV, RV, and LVM can be observed from DA-CA.

{\bf Quantitative Results} 
The quantitative results, as presented in Table.~\ref{tab:cardiac_table}, further highlight the superiority of our approach.
Among the traditional methods, B-spline achieves the highest Dice score, outperforming both SyN and Demons.
Nevertheless, all these methods fall short when compared to neural network-based techniques.
Among the latter, our method significantly surpasses VM ($p<<0.05$ in a paired T-test) in terms of Dice score (both average Dice and Dice of separate regions) and HD.
This results in a 4.4\% improvement in average Dice and an 8.2\% improvement in Dice of LVM.
Neural networks often struggle to learn explicit distance information \cite{liu2018intriguing,islam2020much,zhang2023spatially}, whereas our DA-CA explicitly propagates the information from the edge of LVM to the central area of LV, leading to significant improvement in LVM registration.
Regarding clinical indices, the LVMM calculated based on the results predicted by our method shows no significant difference ($p=0.44$) from the reference (presented in the row of "Before Reg"), which underscores the effectiveness of our method in preserving anatomical structure post-registration.
For qualitative results from ES to ED and ablation study, please refer to the supplementary materials.

\begin{table*}[!t]
\caption{
Quantitative comparisons of cardiac cine-MR registration results produced by DA-CA and other comparator networks. 
Best performing metrics on Dice score and HD are bolded.
For LVEDV and LVMM, those results with no significant difference to the reference are also highlighted in bold.}
\label{tab:cardiac_table}
\begin{center}
\resizebox{1.\columnwidth}{!}{
\begin{tabular}{ lccccccc }
\hline
\hline
 Model    & Avg. Dice (\%) & LVBP Dice (\%) & LVM Dice (\%) & RV Dice (\%) & HD95 (mm) & LVEDV & LVMM  \\
\hline
Before Reg & $56.00 $ & $58.34$ & $38.31$ & $71.35$ & $9.78$& $114.29$ & $84.04$\\
B-spline \cite{marstal2016simpleelastix} &$74.73$ & $79.59$ & $66.06 $ &$78.55$ & $8.68$ &  $107.31$ & $90.69 $\\
Demons \cite{vercauteren2009diffeomorphic}  &$71.22$ & $75.48$ & $60.81 $ &$77.38$  & $9.17$ & $98.15$ & $89.65 $\\
SyN \cite{avants2011reproducible}  & $66.26 $ & $72.73 $ & $55.81 $ & $70.24 $ & $10.85 $ & $102.21 $ & $87.70 $\\
VM \cite{balakrishnan2019voxelmorph}   & $75.49 $ & $81.77 $ & $66.21 $ & $78.50$ & $8.85 $ & $106.82 $ & $80.79 $\\
DA-CA & $\textbf{78.84} $ & $\textbf{84.45} $ & $\textbf{71.66} $ & $\textbf{80.42} $ & $\textbf{8.35} $ & $108.33 $ & $\textbf{84.99}$\\
\hline
\end{tabular}
}
\end{center}
\vspace{-2ex}
\end{table*}

\section{Conclusions}

In conclusion, this paper presents the Directed Accumulator Grid (DAGrid), an enhancement of the Directed Accumulator (DA), that proves effective for dense pixel-wise mapping tasks. 
Applied to two medical applications, skin lesion segmentation and cardiac image registration, DAGrid outperforms leading methods. 
Networks using DAGrid show a significant reduction in network parameter size and FLOPs, while improving the Dice score and Jaccard Index for skin lesion segmentation. 
In cardiac image registration, we also observed improved registration accuracy. 
We believe the potential of DAGrid extends beyond these applications, suggesting its utility in a broader range of dense prediction tasks in medical imaging and beyond.

\section{Appendix}

The appendix section provides a discussion on the limitations of our study and an extensive exploration of our methodologies. 
It offers further insights into the training process, network architecture, and an ablation study, particularly focusing on the application of DA-PA and DA-CA in skin lesion segmentation and cardiac image registration. 
Moreover, the appendix showcases an extensive range of qualitative results from these tasks, serving to further validate our findings and contribute to the overall understanding of our work's implications.

\subsection{Limitations of the Study}

There are three primary limitations identified in our study.
Firstly, the application scope of DAGrid in this study is confined to skin lesion segmentation and cardiac image registration. 
This covers only a small segment of potential medical imaging applications that require dense predictions. 
Furthermore, the circular accumulator and polar accumulator currently only represent circular geometries. 
There exists a multitude of geometric structures within the field of medical imaging that are yet to be explored and leveraged.

Secondly, our approaches for both skin lesion segmentation and cardiac image registration rely solely on 2D feature maps. 
While this is suitable given that skin lesions are a 2D problem and cardiac image registration, with its thick slice images, is suited to 2D, it leaves room for further exploration in 3D space. 
We aim to undertake more investigations to model geometric structures in 3D space in the future.

Lastly, the DAGrid, similar to convolution or grid sampling, is a fundamental element within a neural network. 
The seamless integration of this element into any network framework is vital, an aspect that our study did not delve into. 
Additionally, we also need to study further on how to connect DAGrid with other transformations such as Radon Transform and Fourier Transform, and devise superior slicing modules.

\subsection{Skin Lesion Segmentation}

In this section, we delve deeper into our process, beginning with the specifics of the training and network architecture. We then present an ablation study, followed by a discussion of qualitative results.

\begin{figure}[!ht]
    \centering
    \includegraphics[width=1\columnwidth]{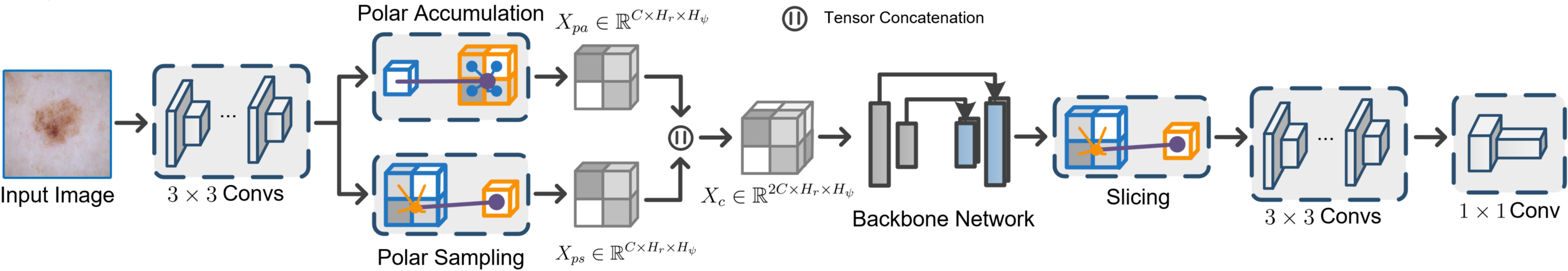}
    \caption{
        The above figure visually illustrates the network architecture for skin lesion segmentation.
        } 
    \label{fig:skin_lesion_network}
\end{figure}

\subsubsection{Implementation Details}

Data augmentation techniques such as flipping were implemented randomly either vertically or horizontally.
Affine transformations were carried out with a variable scale ranging from 0.95 to 1.05, accompanied by a random rotation degree between -5$^{\circ}$ and 5$^{\circ}$.
Furthermore, we employed a motion transformation using a random rotation within the range of -10$^{\circ}$ to 10$^{\circ}$ and random translation of up to 10 pixels.
The resulting transformed images following the application of either affine or motion transformations were obtained through linear interpolation.

We performed all analyses using Python 3.7. 
Our network models, built with the PyTorch library \cite{paszke2019pytorch}, were trained on a machine equipped with two Nvidia Titan XP GPUs.
The Adam optimizer \cite{kingma2014adam} was utilized with an initial learning rate of 0.001, and a multi-step learning rate scheduler set at milestones of 50\%, 70\%, and 90\% of the total epochs, respectively, was employed for network training.
We used a mini-batch size of 24 for training, and the training was stopped after 70 epochs.

\subsubsection{Network Architecture}

We adopted All-Net \cite{zhang2021all} as the backbone network for our skin lesion segmentation tasks. 
To effectively incorporate the polar accumulator (PA) and polar sampling (PS), we deployed three convolution blocks prior to the backbone network. 
Each of these blocks consists of a $3\times 3$ convolution, a subsequent batch normalization \cite{ioffe2015batch}, and a ReLU activation function. 
Following this, either PA, PS, or a combination of both, is applied to the output feature map derived from these blocks. 
The transformed feature map is then fed into the backbone network for further processing. 
The feature map obtained from the backbone network undergoes slicing, which could be either bilinear or parametric. 
This is succeeded by an additional three $3\times 3$ convolution blocks, and eventually, a $1\times1$ convolution operation is performed to generate the logits.
Please refer to Fig.~\ref{fig:skin_lesion_network} for a visual illustration.

\subsubsection{Qualitative Results}

\begin{figure}[!ht]
    \centering
    \includegraphics[width=1\columnwidth]{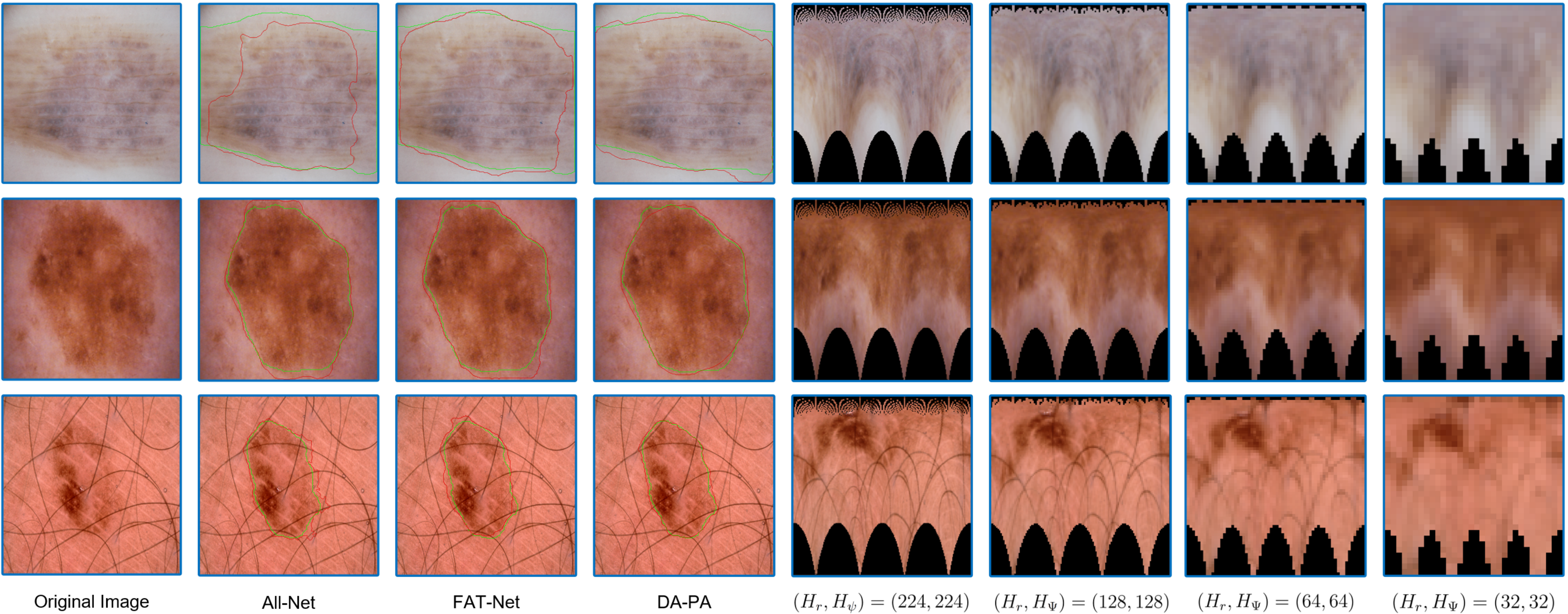}
    \caption{
             A qualitative comparison of skin lesion segmentation performance between our method and other comparative approaches.
             Visualizations of polar accumulation with parameters$(H_r,H_\Psi) = (32,32)$, $(H_r,H_\Psi) = (64,64)$, $(H_r,H_\Psi) = (128,128)$ and $(H_r,H_\Psi) = (224,224)$ are also included.
             The green contour represents the ground-truth segmentation, while the red contour denotes the segmentation output generated by the corresponding network.
        } 
    \label{fig:skin_lesion_results}
\end{figure}

We present qualitative results of DA-PA in comparison with other methods in Fig.~\ref{fig:skin_lesion_results}. 
Visualized results from All-Net and FAT-Net are included, each representing the best CNN and transformer-based methods, respectively. 
Across all three examples, we can observe from the images that the segmentation contour of DA-PA adheres more closely to the ground-truth segmentation contour.

In the top row, both FAT-Net and All-Net under-segment the large lesion, a consequence of an intensity change within the lesion close to its boundary. 
In the middle row, both FAT-Net and All-Net over-segment the lesion. 
This over-segmentation occurs because the appearance of the lesion and background is strikingly similar in this case. 
In the third row, both FAT-Net and All-Net are influenced by the messy hair scattered across and around the lesion, resulting in a non-smooth segmentation boundary.

\subsubsection{Ablation Study}

We demonstrate the efficacy of each component within DA-PA via an ablation study, which is detailed in Table \ref{tab:segmentation_ablation}. 
Through a comparison of the results obtained from models \# 0, \# 1, and \# 2, it is evident that integrating the polar transformation into the neural network improves the skin lesion segmentation, and PA with bilinear sampling kernel outperforms PS. 
Intriguingly, further performance enhancement is achieved by parametrizing the slicing process of DA-PA with a learnable linear combination module, as evinced by a comparison between models \# 2 and \# 3.

Given that both PS and PA transform feature maps into the polar representation in distinct ways, it seems logical to fuse them. 
Nevertheless, when PS and PA are fused at a high sampling rate with $(H_r,H_\Psi) = (224,224)$, as in model \# 4, there is a degradation in performance compared to the counterpart without fusion (model \# 3). 
Interestingly, maintaining a relatively low sampling rate for fusion with $(H_r,H_\Psi) = (64,64)$, as in model \# 6, the performance is on par with model \# 7.
This contradiction suggests that when applied at a high sampling rate, PS introduces redundant information at the center of the image and overlooks details distant from the center.
Consequently, concatenating PS and PA through the channel dimension disrupts the overall feature map, leading to a decrease in performance. 
However, when applied at a low sampling rate, the redundancy is minimized and fewer details are lost because there is not as much information to begin with, resulting in a balanced performance.

Given that computing an additional PS requires virtually no additional computational effort, we opted to use the fusion setting for the rest of the models with varying sampling rates, namely $(H_r,H_\Psi) = (128,128)$, and $(H_r,H_\Psi) = (32,32)$. 
As seen from models \# 4, \# 8, \# 6, and \# 5, the performance initially increases, peaking when $(H_r,H_\Psi) = (64,64)$, and subsequently starts to decline in model \# 5 with $(H_r,H_\Psi) = (32,32)$.
The network processing polar representations exhibits a desirable property of being equivariant to both rotation and scale. 
Consequently, as the resolution of the polar representation decreases, the receptive field centered on the lesion enlarges, capturing a more comprehensive contextual understanding, which proves advantageous for skin lesion segmentation. 
Moreover, a smaller polar representation leads to larger smooth segments of the image sliced back from the polar representation, which aids in reducing noise surrounding the lesion. 
However, when the resolution becomes too small, the information loss cannot be adequately compensated for, even with parametric sampling, resulting in a degradation in performance.

\begin{table*}[!ht]
\caption{
    Ablation study of our proposed method on skin lesion segmentation. 
    In the ``Transformation'' column, 'PS+PA' denotes the fusion of PS and PA representations through the channel dimension. 
    It's important to note that all slicing related to PS is performed using bilinear interpolation; the ``Slicing Type'' column only refers to the slicing methods used in DA-PA. 
    The '$\times$' symbol indicates that the operation associated with that column has been omitted.
}
\label{tab:segmentation_ablation}
\begin{center}
\resizebox{1.\columnwidth}{!}{
\begin{tabular}{ lccccccccc }
\hline
\hline
 Model & Transformation & Polar Size & Slicing Type & Avg. Dice (\%) & Avg. Precision (\%) & Avg. Sensitivity (\%) & Avg. Jaccard Index (\%)  \\
\hline
\# 0 & $\times$ & $\times$ & $\times$ & 86.16 & 88.54 & 88.71 & 78.04 \\
\# 1 & PS & $(H_r=224,H_\psi=224)$ & Bilinear & 87.58 & 88.01 & 90.53 & 79.33\\
\# 2 & PA & $(H_r=224,H_\psi=224)$ & Bilinear & 88.23  & 90.23 & 89.92 & 80.76\\
\# 3 & PA & $(H_r=224,H_\psi=224)$ & Parametric & 88.62 & 89.93 & 90.78 & 81.26\\
\# 4 & PS+PA & $(H_r=224,H_\psi=224)$ & Parametric & 88.01   & 89.80 &  90.05 & 80.44 \\
\# 5 & PS+PA & $(H_r=32,H_\psi=32)$ & Parametric & 89.37 & 90.30 & 91.73 & 82.46 \\
\# 6 & PS+PA & $(H_r=64,H_\psi=64)$ & Parametric & 89.41 & 90.11 & 92.08 & 82.56 \\
\# 7 & PA & $(H_r=64,H_\psi=64)$ & Parametric & 89.35 & 88.98 & 92.90 & 82.30\\
\# 8 & PS+PA & $(H_r=128,H_\psi=128)$ & Parametric & 88.85 & 91.68 & 89.90 & 82.02 \\
\hline
\end{tabular}
}
\end{center}
\end{table*}

\subsection{Cardiac Image Registration}

In this section, we elaborate further on the specifics of network training, the architecture of the network, and qualitative results from End-Systole (ES) to End-Diastole (ED). 
Additionally, we present an in-depth ablation study to investigate the impact of different components and parameters.

\subsubsection{Implementation Details}

We augmented our ACDC datasets for cardiac image registration with random flipping in the coronal and sagittal directions.
All analyses were performed using Python 3.7, and our network models, along with the comparators, were constructed using the PyTorch library \cite{paszke2019pytorch}.
We trained these models on a machine equipped with two Nvidia Titan XP GPUs.
To evaluate the similarity between the registered moving image and the fixed image, we used the Normalized Cross Correlation (NCC) loss with a window size of $(15,15,3)$ (reflecting the voxel size of $1.8\times 1.8 \times 10.0$).
This window size reflects the image spacing of $1.8\times 1.8 \times 10.0$.
To promote smoothness, we employed the L1 norm of the gradient of the registration field.
These two metrics were applied in a ratio of 1:0.01.
For network training, we utilized the Adam optimizer \cite{kingma2014adam} with an initial learning rate of $1e-3$.
Furthermore, we employed a polynomial learning rate scheduler with a decay rate of 0.9.
The training was conducted with a mini-batch size of 4 and was terminated after 400 epochs.

\begin{figure}[!ht]
    \centering
    \includegraphics[width=1\columnwidth]{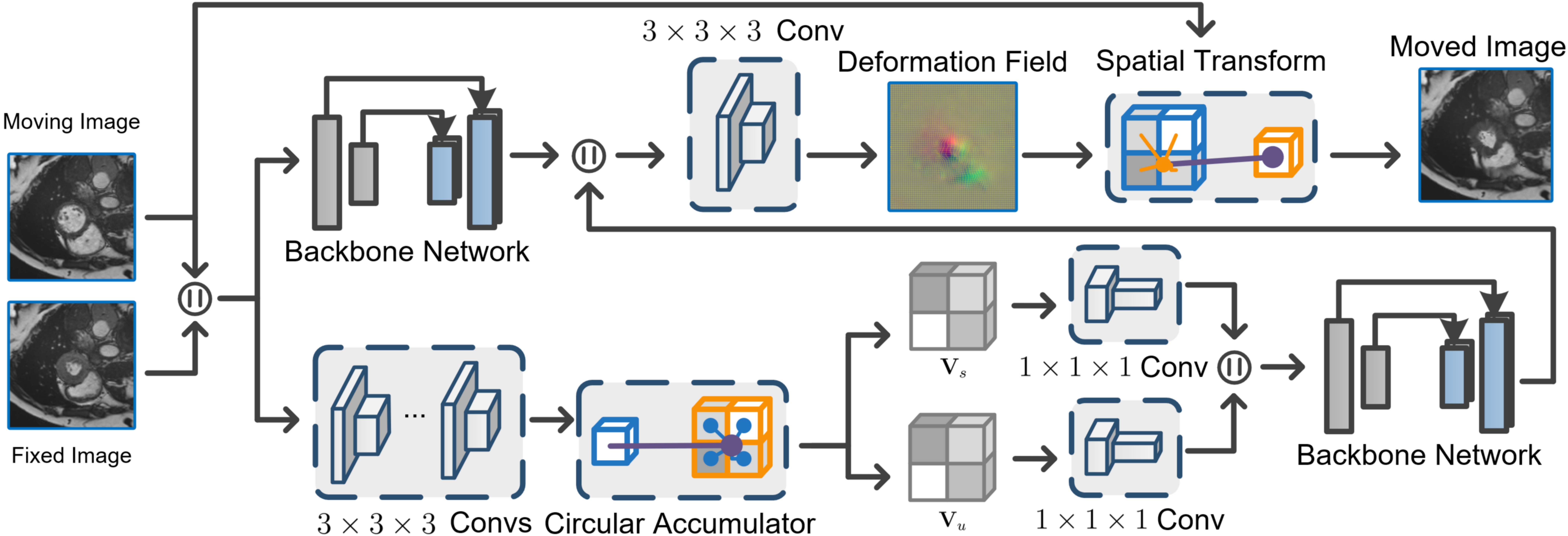}
    \caption{
        The above figure visually illustrates the network architecture for cardiac image registration.
        } 
    \label{fig:cardiac_network}
\end{figure}

\subsubsection{Network Architecture}

We employed VoxelMorph (VM) \cite{zhang2021all} as the backbone network for our cardiac image registration task. 
To integrate the circular accumulator (CA) effectively, we utilized a dual-branch network. 
One branch mirrors the VM structure, while the other branch is dedicated to CA. 
Importantly, these branches operate independently with their own encoder-decoder networks.
For the CA branch, we deployed five convolution blocks prior to CA application. 
Each of these blocks consists of a $3\times 3 \times 3$ convolution, followed by batch normalization \cite{ioffe2015batch}, and a ReLU activation function.
Applying CA to the output feature map of these convolutions, we obtain $\mathbf{V}_s$ and $\mathbf{V}_u$ with different radius ranges denoted by $N$. 
To embed features more effectively, these vectors undergo a separate $1\times 1 \times 1$ convolution block.
Next, they are concatenated along the channel dimension and fed into the respective backbone network. 
The feature maps generated by the two backbone networks are further concatenated and passed through a $3\times 3 \times 3$ convolution block to yield the deformation field. 
For a visual illustration of this process, please refer to Fig.~\ref{fig:cardiac_network}.

\subsubsection{Qualitative Results}

\begin{figure}[!t]
    \centering
    \includegraphics[width=1\columnwidth]{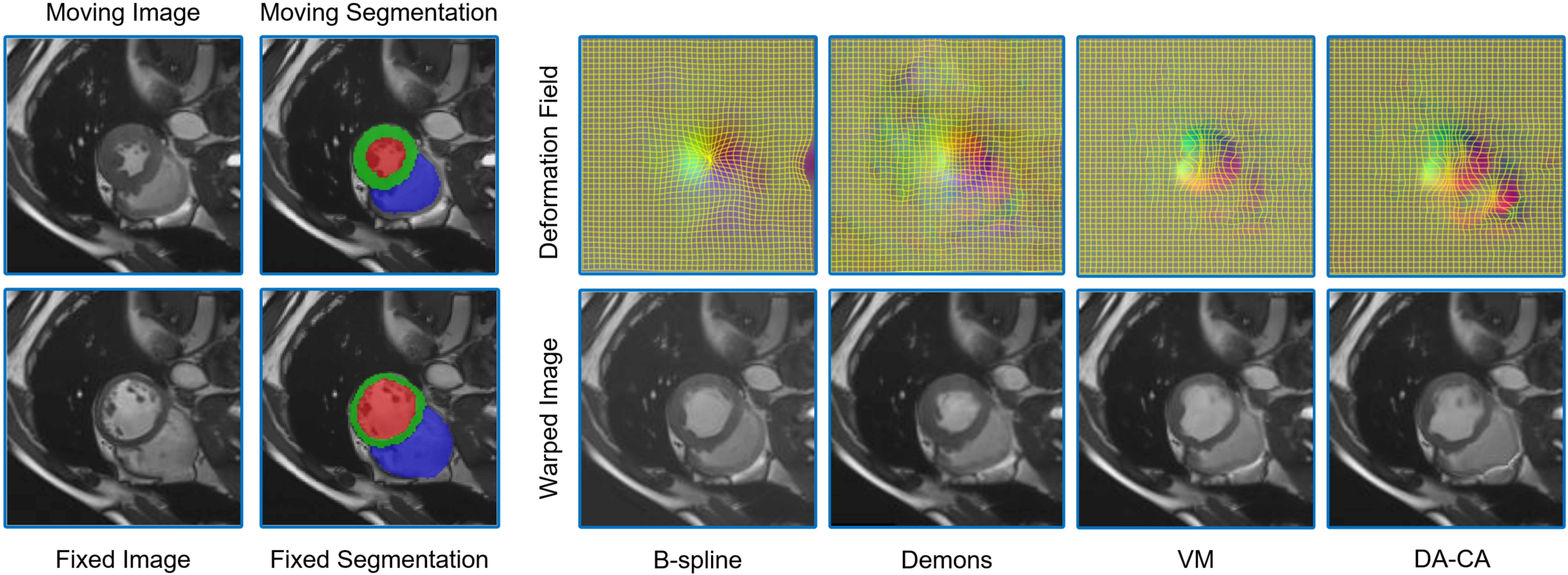}
    \caption{
        A qualitative comparison of cardiac image registration performance, showcasing our method alongside other comparative approaches. 
        The moving image corresponds to the end-systolic phase (ES), while the fixed image corresponds to the end-diastolic phase (ED).
        } 
    \label{fig:registration_es_ed}
\end{figure}

We present additional qualitative results of DA-CA in comparison with other methods in Fig.~\ref{fig:registration_es_ed}, where the moving image corresponds to the end-systolic phase (ES), while the fixed image corresponds to the end-diastolic phase (ED). 
When registering from ED to ES, the left ventricular mass undergoes significant distortion, while when registering from ES to ED, the substantial distortion occurs in the left ventricle. 
This makes registration from ED to ES more challenging than from ES to ED.

In ED to ES registration, the network must generate precise deformation fields for the left ventricular mass that points from the interior to the edges. 
However, for ES to ED, the network simply needs to generate a deformation field that points from the exterior towards the center of the left ventricle and its surroundings. 
Even with some small deviations in this context, the overall impact is negligible. 
However, for ED to ES, due to the small proportion of left ventricular mass in the ED phase, even the slight deviation can lead to a failure in registration.

As seen in Fig.\ref{fig:registration_es_ed}, the results from other methods seem more satisfactory than those in ED to ES due to the aforementioned analysis. 
Nonetheless, it is still apparent from the figure that our method, which explicitly leverages long-range radial information, yields the most distinct deformation field, especially for the right ventricle.

\subsubsection{Ablation Study}

\begin{table*}[!ht]
\caption{Ablation study of our proposed method on cardiac cine-MR registration. For each model, the check mark indicates the characteristic is present.}
\label{tab:registration_ablation}
\begin{center}
\resizebox{1.\columnwidth}{!}{
\begin{tabular}{ lccccccccc }
\hline
\hline
 Model &  Symmetric & Sampling & \# of Branches & $N$  & Avg. Dice (\%) & LVBP Dice (\%) & LVM Dice (\%) & RV Dice (\%) & HD95 (mm) \\
\hline
\# 0 & $\checkmark$ & Bilinear & 2 & $\{5\}$ & $77.67$ & $83.18$    & $70.35$ & $79.47$ & $8.52$ \\
\# 1 & $\checkmark$ & Bilinear & 2 & $\{10,5\}$ &$78.35$ & $83.72$   & $71.25$ & $80.07$ & $8.38$ \\
\# 2 & $\checkmark$ & Bilinear & 2 & $\{15,10\}$ & $76.94$ & $82.89$ & $69.36$ & $78.56$ & $8.60$ \\
\# 3 & $\checkmark$ & Bilinear & 2 & $\{15,10,5\}$ & $78.84$ & $84.45$  & $71.66$ & $80.42$ & $8.35$ \\
\# 4 & $\checkmark$ & Nearest & 2 & $\{15,10,5\}$ & $78.18$ & $83.60$  & $70.92$ & $80.02$ & $8.47$ \\
\# 5 & $\checkmark$ & Bilinear & 1 & $\{15,10,5\}$ & $78.76$ & $84.37$   & $71.84$ & $80.06$ & $8.48$ \\
\# 6 & $\times$ & Bilinear & 2 & $\{15,10,5\}$ & $74.69$ & $80.35$    & $66.14$ & $77.57$ & $8.66$ \\
\# 7 \cite{chen2022joint} & $\times$ & $\times$ & 1 & $\times$ & $76.52$ & $82.06$    & $68.44$ & $79.06$ & $8.73$ \\
\hline
\end{tabular}
}
\end{center}
\end{table*}

We demonstrate the effectiveness of each component within DA-CA via an ablation study, detailed in Table \ref{tab:registration_ablation}. 
By comparing the results obtained from models \# 0, \# 1, \# 2, and \# 3, it becomes clear that integrating the circular accumulator with different radius ranges into the neural network improves cardiac image registration performance.
In addition, it shows that the small radius perserves more useful information than large radius.

In comparing models \# 3 and \# 4, we find that bilinear sampling with backpropagated gradients for sampling grids performs better than nearest sampling without gradients for the grids. 
When contrasting models \# 3 and \# 5, we find that a network with a single branch using CA performs very well, but given the small network size of VM, adding another branch for processing raw images is our preferred choice.

Comparing \# 3 and \# 6, we discern that the symmetric information is critically important for cardiac image registration. 
This is because at times, if the gradient of the feature map points outward from the ventricle, the long-range information is entirely lost. 
Therefore, by including symmetric information, we can capture this long-range information regardless of its direction.

Lastly, we compared our method with the co-attention based registration network \cite{chen2022joint}, which uses co-attention to realize implicit long-range information propagation. 
By comparing models \# 3 and \# 7, we find that both methods outperform the original VM network, which lacks long-range information. 
However, our DA-CA, with explicit long-range information propagation along the geometric structure, outperforms the co-attention network.

\begin{figure}[!t]
    \centering
    \includegraphics[width=1\columnwidth]{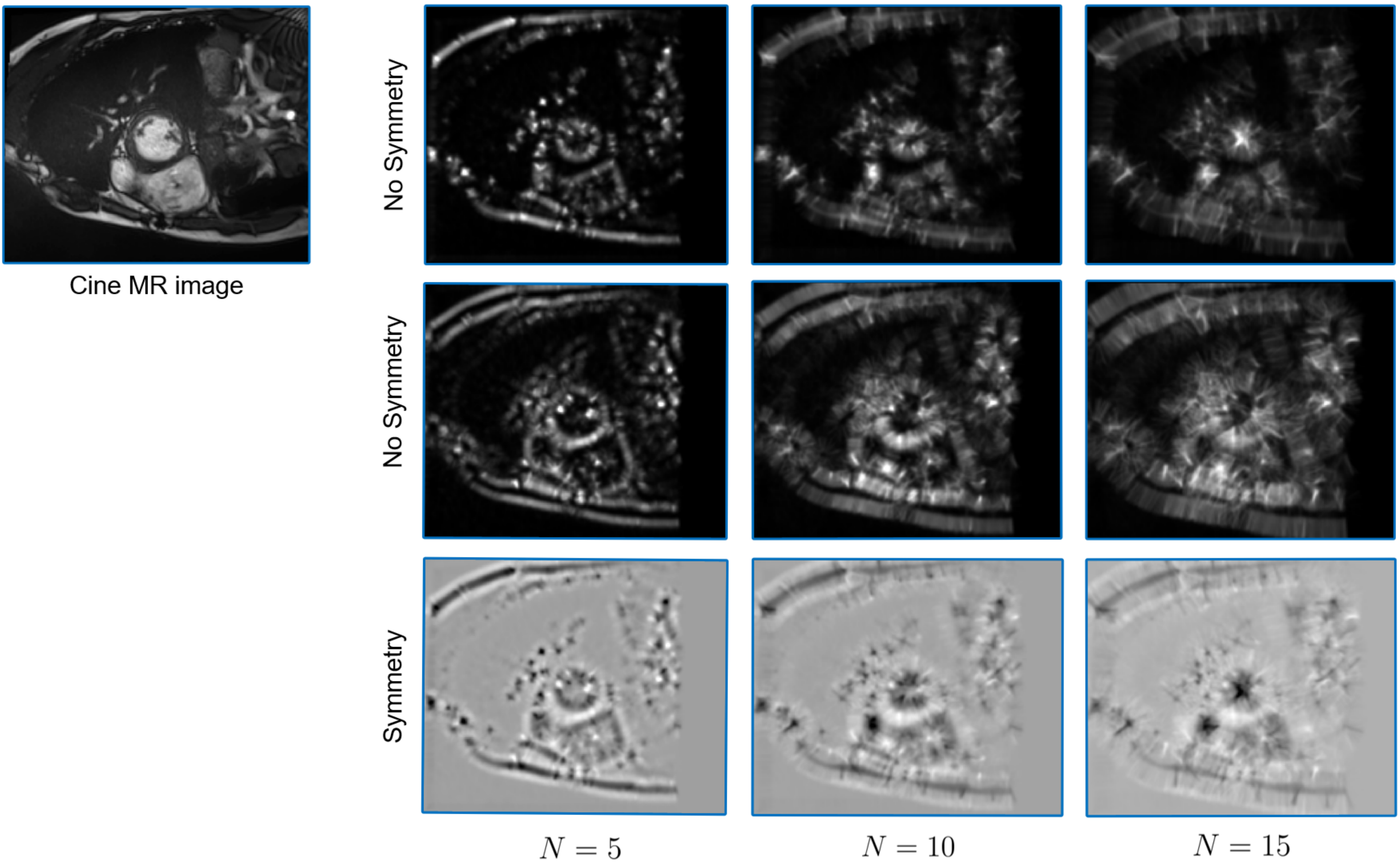}
    \caption{
        A visual illustration of the circular accumulator with different parameters. The importance of symmetric information is demonstrated in the first two rows, where opposite gradient directions have been manually created to showcase the impact of this element.
        } 
    \label{fig:hct_examples}
\end{figure}

\subsubsection{Visualization of the CA Transformed Representation}

Fig. \ref{fig:hct_examples} demonstrates the CA-transformed representation with various parameters, specifically focusing on the radius range $N$ and the inclusion of symmetric information. 
This exploration is rooted in our concern that without considering symmetry, the gradient direction may inadvertently point outward from the ventricle, leading to a potential loss of information.

To investigate this, we manually created two cases, illustrated in the first two rows of Fig. \ref{fig:hct_examples}. 
Here, the sampling grids, represented by image gradients, have opposite directions. 
Let $\mathcal{G}_1 = {\mathbf{G}[k]=(\mathbf{G}^x[k], \mathbf{G}^y[k])~|k \in \mathbb{Z}^+, 1 \leq k \leq N }$ be the sampling grids of the first row. 
The sampling grids for the second row then mirror those of the first row, following $\mathcal{G}_2 = {\mathbf{G}[k]=(-\mathbf{G}^x[k], -\mathbf{G}^y[k])|~k \in \mathbb{Z}^+, 1 \leq k \leq N }$.

As observed in Fig. \ref{fig:hct_examples}, when the information points to the correct direction, it converges towards the ventricle. 
In contrast, when pointing in the incorrect direction, the information tends to disperse outwards. 
By employing a symmetric formulation, as illustrated in the third row of Fig. \ref{fig:hct_examples}, we can avoid such errors. 
This is particularly crucial because during convolution, the neural network is unaware of the ultimate direction of the gradients.

\newpage

\printbibliography

@inproceedings{zhang2021efficient,
  title={Efficient folded attention for medical image reconstruction and segmentation},
  author={Zhang, Hang and Zhang, Jinwei and Wang, Rongguang and Zhang, Qihao and Spincemaille, Pascal and Nguyen, Thanh D and Wang, Yi},
  booktitle={Proceedings of the AAAI Conference on Artificial Intelligence},
  volume={35},
  number={12},
  pages={10868--10876},
  year={2021}
}

@article{bernard2018deep,
  title={{Deep Learning Techniques for Automatic MRI Cardiac Multi-structures Segmentation and Diagnosis: Is the Problem Solved?}},
  author={Bernard, Olivier and Lalande, Alain and Zotti, Clement and Cervenansky, Frederick and Yang, Xin and Heng, Pheng-Ann and Cetin, Irem and Lekadir, Karim and Camara, Oscar and Ballester, Miguel Angel Gonzalez and others},
  journal={IEEE Transactions on Medical Imaging},
  volume={37},
  number={11},
  pages={2514--2525},
  year={2018},
  publisher={ieee}
}

@article{avants2011reproducible,
  title={{A Reproducible Evaluation of ANTs Similarity Metric Performance in Brain Image Registration}},
  author={Avants, Brian B and Tustison, Nicholas J and Song, Gang and Cook, Philip A and Klein, Arno and Gee, James C},
  journal={Neuroimage},
  volume={54},
  number={3},
  pages={2033--2044},
  year={2011},
  publisher={Elsevier}
}

@inproceedings{marstal2016simpleelastix,
  title={{SimpleElastix: A User-friendly, Multi-lingual Library for Medical Image Registration}},
  author={Marstal, Kasper and Berendsen, Floris and Staring, Marius and Klein, Stefan},
  booktitle={Proceedings of the IEEE Conference on Computer Vision and Pattern Recognition Workshops},
  pages={134--142},
  year={2016}
}

@article{zhang2021all,
  title={ALL-Net: Anatomical information lesion-wise loss function integrated into neural network for multiple sclerosis lesion segmentation},
  author={Zhang, Hang and Zhang, Jinwei and Li, Chao and Sweeney, Elizabeth M and Spincemaille, Pascal and Nguyen, Thanh D and Gauthier, Susan A and Wang, Yi and Marcille, Melanie},
  journal={NeuroImage: Clinical},
  volume={32},
  pages={102854},
  year={2021},
  publisher={Elsevier}
}

@article{zhang2022qsmrim,
  title={QSMRim-Net: Imbalance-aware learning for identification of chronic active multiple sclerosis lesions on quantitative susceptibility maps},
  author={Zhang, Hang and Nguyen, Thanh D and Zhang, Jinwei and Marcille, Melanie and Spincemaille, Pascal and Wang, Yi and Gauthier, Susan A and Sweeney, Elizabeth M},
  journal={NeuroImage: Clinical},
  volume={34},
  pages={102979},
  year={2022},
  publisher={Elsevier}
}

@article{chen2007real,
  title={Real-time edge-aware image processing with the bilateral grid},
  author={Chen, Jiawen and Paris, Sylvain and Durand, Fr{\'e}do},
  journal={ACM Transactions on Graphics (TOG)},
  volume={26},
  number={3},
  pages={103--es},
  year={2007},
  publisher={Acm New York, NY, USA}
}

@article{gharbi2017deep,
  title={Deep bilateral learning for real-time image enhancement},
  author={Gharbi, Micha{\"e}l and Chen, Jiawen and Barron, Jonathan T and Hasinoff, Samuel W and Durand, Fr{\'e}do},
  journal={ACM Transactions on Graphics (TOG)},
  volume={36},
  number={4},
  pages={1--12},
  year={2017},
  publisher={ACM New York, NY, USA}
}

@inproceedings{esteves2018polar,
  title={Polar Transformer Networks},
  author={Esteves, Carlos and Allen-Blanchette, Christine and Zhou, Xiaowei and Daniilidis, Kostas},
  booktitle={International Conference on Learning Representations},
  year={2018}
}

@article{balakrishnan2019voxelmorph,
  title={VoxelMorph: a learning framework for deformable medical image registration},
  author={Balakrishnan, Guha and Zhao, Amy and Sabuncu, Mert R and Guttag, John and Dalca, Adrian V},
  journal={IEEE transactions on medical imaging},
  volume={38},
  number={8},
  pages={1788--1800},
  year={2019},
  publisher={IEEE}
}

@inproceedings{liu2021swin,
  title={Swin transformer: Hierarchical vision transformer using shifted windows},
  author={Liu, Ze and Lin, Yutong and Cao, Yue and Hu, Han and Wei, Yixuan and Zhang, Zheng and Lin, Stephen and Guo, Baining},
  booktitle={Proceedings of the IEEE/CVF International Conference on Computer Vision},
  pages={10012--10022},
  year={2021}
}

@inproceedings{dosovitskiy2020image,
  title={An Image is Worth 16x16 Words: Transformers for Image Recognition at Scale},
  author={Dosovitskiy, Alexey and Beyer, Lucas and Kolesnikov, Alexander and Weissenborn, Dirk and Zhai, Xiaohua and Unterthiner, Thomas and Dehghani, Mostafa and Minderer, Matthias and Heigold, Georg and Gelly, Sylvain and others},
  booktitle={International Conference on Learning Representations},
  year={2020}
}

@inproceedings{matsoukas2022makes,
  title={What Makes Transfer Learning Work For Medical Images: Feature Reuse \& Other Factors},
  author={Matsoukas, Christos and Haslum, Johan Fredin and Sorkhei, Moein and S{\"o}derberg, Magnus and Smith, Kevin},
  booktitle={Proceedings of the IEEE/CVF Conference on Computer Vision and Pattern Recognition},
  pages={9225--9234},
  year={2022}
}

@article{osti4746348,
title = {METHOD AND MEANS FOR RECOGNIZING COMPLEX PATTERNS},
author = {Hough, P V.C.},
doi = {},
url = {https://www.osti.gov/biblio/4746348}, journal = {},
place = {United States},
year = {1962},
month = {12}
}

@article{milletari2017hough,
  title={Hough-CNN: deep learning for segmentation of deep brain regions in MRI and ultrasound},
  author={Milletari, Fausto and Ahmadi, Seyed-Ahmad and Kroll, Christine and Plate, Annika and Rozanski, Verena and Maiostre, Juliana and Levin, Johannes and Dietrich, Olaf and Ertl-Wagner, Birgit and B{\"o}tzel, Kai and others},
  journal={Computer Vision and Image Understanding},
  volume={164},
  pages={92--102},
  year={2017},
  publisher={Elsevier}
}

@inproceedings{qi2019deep,
  title={Deep hough voting for 3d object detection in point clouds},
  author={Qi, Charles R and Litany, Or and He, Kaiming and Guibas, Leonidas J},
  booktitle={proceedings of the IEEE/CVF International Conference on Computer Vision},
  pages={9277--9286},
  year={2019}
}

@inproceedings{qi2020imvotenet,
  title={Imvotenet: Boosting 3d object detection in point clouds with image votes},
  author={Qi, Charles R and Chen, Xinlei and Litany, Or and Guibas, Leonidas J},
  booktitle={Proceedings of the IEEE/CVF conference on computer vision and pattern recognition},
  pages={4404--4413},
  year={2020}
}

@article{zhang2021memory,
  title={Memory U-Net: Memorizing Where to Vote for Lesion Instance Segmentation},
  author={Zhang, Hang and Zhang, Jinwei and Yang, Gufeng and Spincemaille, Pascal and Nguyen, Thanh D and Wang, Yi},
  year={2021}
}

@inproceedings{lin2020deep,
  title={Deep hough-transform line priors},
  author={Lin, Yancong and Pintea, Silvia L and Gemert, Jan C van},
  booktitle={European Conference on Computer Vision},
  pages={323--340},
  year={2020},
  organization={Springer}
}

@article{zhao2021deep,
  title={Deep hough transform for semantic line detection},
  author={Zhao, Kai and Han, Qi and Zhang, Chang-Bin and Xu, Jun and Cheng, Ming-Ming},
  journal={IEEE Transactions on Pattern Analysis and Machine Intelligence},
  year={2021},
  publisher={IEEE}
}

@inproceedings{xie2015deep,
  title={Deep voting: A robust approach toward nucleus localization in microscopy images},
  author={Xie, Yuanpu and Kong, Xiangfei and Xing, Fuyong and Liu, Fujun and Su, Hai and Yang, Lin},
  booktitle={International Conference on Medical Image Computing and Computer-Assisted Intervention},
  pages={374--382},
  year={2015},
  organization={Springer}
}

@inproceedings{min2019hyperpixel,
  title={Hyperpixel flow: Semantic correspondence with multi-layer neural features},
  author={Min, Juhong and Lee, Jongmin and Ponce, Jean and Cho, Minsu},
  booktitle={Proceedings of the IEEE/CVF International Conference on Computer Vision},
  pages={3395--3404},
  year={2019}
}

@inproceedings{lee2021deep,
  title={Deep hough voting for robust global registration},
  author={Lee, Junha and Kim, Seungwook and Cho, Minsu and Park, Jaesik},
  booktitle={Proceedings of the IEEE/CVF International Conference on Computer Vision},
  pages={15994--16003},
  year={2021}
}

@inproceedings{zhao20223d,
  title={3D Room Layout Estimation from a Cubemap of Panorama Image via Deep Manhattan Hough Transform},
  author={Zhao, Yining and Wen, Chao and Xue, Zhou and Gao, Yue},
  booktitle={European Conference on Computer Vision},
  pages={637--654},
  year={2022},
  organization={Springer}
}

@inproceedings{deng2009imagenet,
  title={Imagenet: A large-scale hierarchical image database},
  author={Deng, Jia and Dong, Wei and Socher, Richard and Li, Li-Jia and Li, Kai and Fei-Fei, Li},
  booktitle={2009 IEEE conference on computer vision and pattern recognition},
  pages={248--255},
  year={2009},
  organization={Ieee}
}

@inproceedings{lin2014microsoft,
  title={Microsoft coco: Common objects in context},
  author={Lin, Tsung-Yi and Maire, Michael and Belongie, Serge and Hays, James and Perona, Pietro and Ramanan, Deva and Doll{\'a}r, Piotr and Zitnick, C Lawrence},
  booktitle={European conference on computer vision},
  pages={740--755},
  year={2014},
  organization={Springer}
}

@inproceedings{kervadec2019boundary,
  title={Boundary loss for highly unbalanced segmentation},
  author={Kervadec, Hoel and Bouchtiba, Jihene and Desrosiers, Christian and Granger, Eric and Dolz, Jose and Ayed, Ismail Ben},
  booktitle={International conference on medical imaging with deep learning},
  pages={285--296},
  year={2019},
  organization={PMLR}
}

@article{karimi2019reducing,
  title={Reducing the hausdorff distance in medical image segmentation with convolutional neural networks},
  author={Karimi, Davood and Salcudean, Septimiu E},
  journal={IEEE Transactions on medical imaging},
  volume={39},
  number={2},
  pages={499--513},
  year={2019},
  publisher={IEEE}
}

@INPROCEEDINGS{9434085,  author={Zhang, Hang and Zhang, Jinwei and Wang, Rongguang and Zhang, Qihao and Gauthier, Susan A. and Spincemaille, Pascal and Nguyen, Thanh D. and Wang, Yi},  booktitle={2021 IEEE 18th International Symposium on Biomedical Imaging (ISBI)},   title={Geometric Loss For Deep Multiple Sclerosis Lesion Segmentation},   year={2021},  volume={},  number={},  pages={24-28},  doi={10.1109/ISBI48211.2021.9434085}}

@inproceedings{ma2020distance,
  title={How distance transform maps boost segmentation CNNs: an empirical study},
  author={Ma, Jun and Wei, Zhan and Zhang, Yiwen and Wang, Yixin and Lv, Rongfei and Zhu, Cheng and Gaoxiang, Chen and Liu, Jianan and Peng, Chao and Wang, Lei and others},
  booktitle={Medical Imaging with Deep Learning},
  pages={479--492},
  year={2020},
  organization={PMLR}
}

@article{teng2020accumulated,
  title={Accumulated polar feature-based deep learning for efficient and lightweight automatic modulation classification with channel compensation mechanism},
  author={Teng, Chieh-Fang and Chou, Ching-Yao and Chen, Chun-Hsiang and Wu, An-Yeu},
  journal={IEEE Transactions on Vehicular Technology},
  volume={69},
  number={12},
  pages={15472--15485},
  year={2020},
  publisher={IEEE}
}

@inproceedings{xie2020polarmask,
  title={Polarmask: Single shot instance segmentation with polar representation},
  author={Xie, Enze and Sun, Peize and Song, Xiaoge and Wang, Wenhai and Liu, Xuebo and Liang, Ding and Shen, Chunhua and Luo, Ping},
  booktitle={Proceedings of the IEEE/CVF conference on computer vision and pattern recognition},
  pages={12193--12202},
  year={2020}
}

@inproceedings{xu2019explicit,
  title={Explicit shape encoding for real-time instance segmentation},
  author={Xu, Wenqiang and Wang, Haiyang and Qi, Fubo and Lu, Cewu},
  booktitle={Proceedings of the IEEE/CVF International Conference on Computer Vision},
  pages={5168--5177},
  year={2019}
}

@inproceedings{park2022eigencontours,
  title={Eigencontours: Novel contour descriptors based on low-rank approximation},
  author={Park, Wonhui and Jin, Dongkwon and Kim, Chang-Su},
  booktitle={Proceedings of the IEEE/CVF Conference on Computer Vision and Pattern Recognition},
  pages={2667--2675},
  year={2022}
}

@inproceedings{ebel2019beyond,
  title={Beyond cartesian representations for local descriptors},
  author={Ebel, Patrick and Mishchuk, Anastasiia and Yi, Kwang Moo and Fua, Pascal and Trulls, Eduard},
  booktitle={Proceedings of the IEEE/CVF international conference on computer vision},
  pages={253--262},
  year={2019}
}

@inproceedings{schmidt2018cell,
  title={Cell detection with star-convex polygons},
  author={Schmidt, Uwe and Weigert, Martin and Broaddus, Coleman and Myers, Gene},
  booktitle={International Conference on Medical Image Computing and Computer-Assisted Intervention},
  pages={265--273},
  year={2018},
  organization={Springer}
}

@article{stringer2021cellpose,
  title={Cellpose: a generalist algorithm for cellular segmentation},
  author={Stringer, Carsen and Wang, Tim and Michaelos, Michalis and Pachitariu, Marius},
  journal={Nature methods},
  volume={18},
  number={1},
  pages={100--106},
  year={2021},
  publisher={Nature Publishing Group}
}

@article{liu2018intriguing,
  title={An intriguing failing of convolutional neural networks and the coordconv solution},
  author={Liu, Rosanne and Lehman, Joel and Molino, Piero and Petroski Such, Felipe and Frank, Eric and Sergeev, Alex and Yosinski, Jason},
  journal={Advances in neural information processing systems},
  volume={31},
  year={2018}
}

@article{zhang2020fidelity,
  title={Fidelity imposed network edit (FINE) for solving ill-posed image reconstruction},
  author={Zhang, Jinwei and Liu, Zhe and Zhang, Shun and Zhang, Hang and Spincemaille, Pascal and Nguyen, Thanh D and Sabuncu, Mert R and Wang, Yi},
  journal={Neuroimage},
  volume={211},
  pages={116579},
  year={2020},
  publisher={Elsevier}
}

@article{jaderberg2015spatial,
  title={Spatial transformer networks},
  author={Jaderberg, Max and Simonyan, Karen and Zisserman, Andrew and others},
  journal={Advances in neural information processing systems},
  volume={28},
  year={2015}
}

@inproceedings{he2016deep,
  title={Deep residual learning for image recognition},
  author={He, Kaiming and Zhang, Xiangyu and Ren, Shaoqing and Sun, Jian},
  booktitle={Proceedings of the IEEE conference on computer vision and pattern recognition},
  pages={770--778},
  year={2016}
}

@inproceedings{ioffe2015batch,
  title={Batch normalization: Accelerating deep network training by reducing internal covariate shift},
  author={Ioffe, Sergey and Szegedy, Christian},
  booktitle={International conference on machine learning},
  pages={448--456},
  year={2015},
  organization={PMLR}
}

@article{simonyan2014very,
  title={Very deep convolutional networks for large-scale image recognition},
  author={Simonyan, Karen and Zisserman, Andrew},
  journal={arXiv preprint arXiv:1409.1556},
  year={2014}
}

@article{paszke2019pytorch,
  title={Pytorch: An imperative style, high-performance deep learning library},
  author={Paszke, Adam and Gross, Sam and Massa, Francisco and Lerer, Adam and Bradbury, James and Chanan, Gregory and Killeen, Trevor and Lin, Zeming and Gimelshein, Natalia and Antiga, Luca and others},
  journal={Advances in neural information processing systems},
  volume={32},
  year={2019}
}

@article{kingma2014adam,
  title={Adam: A method for stochastic optimization},
  author={Kingma, Diederik P and Ba, Jimmy},
  journal={arXiv preprint arXiv:1412.6980},
  year={2014}
}

@inproceedings{wang2018non,
  title={Non-local neural networks},
  author={Wang, Xiaolong and Girshick, Ross and Gupta, Abhinav and He, Kaiming},
  booktitle={Proceedings of the IEEE conference on computer vision and pattern recognition},
  pages={7794--7803},
  year={2018}
}

@inproceedings{xu2021bilateral,
  title={Bilateral grid learning for stereo matching networks},
  author={Xu, Bin and Xu, Yuhua and Yang, Xiaoli and Jia, Wei and Guo, Yulan},
  booktitle={Proceedings of the IEEE/CVF Conference on Computer Vision and Pattern Recognition},
  pages={12497--12506},
  year={2021}
}

@inproceedings{tomasi1998bilateral,
  title={Bilateral filtering for gray and color images},
  author={Tomasi, Carlo and Manduchi, Roberto},
  booktitle={Sixth international conference on computer vision (IEEE Cat. No. 98CH36271)},
  pages={839--846},
  year={1998},
  organization={IEEE}
}

@inproceedings{chen2021deep,
  title={A deep discontinuity-preserving image registration network},
  author={Chen, Xiang and Xia, Yan and Ravikumar, Nishant and Frangi, Alejandro F},
  booktitle={Medical Image Computing and Computer Assisted Intervention--MICCAI 2021: 24th International Conference, Strasbourg, France, September 27--October 1, 2021, Proceedings, Part IV 24},
  pages={46--55},
  year={2021},
  organization={Springer}
}

@article{wu2022fat,
  title={FAT-Net: Feature adaptive transformers for automated skin lesion segmentation},
  author={Wu, Huisi and Chen, Shihuai and Chen, Guilian and Wang, Wei and Lei, Baiying and Wen, Zhenkun},
  journal={Medical image analysis},
  volume={76},
  pages={102327},
  year={2022},
  publisher={Elsevier}
}

@article{aggarwal2018modl,
  title={MoDL: Model-based deep learning architecture for inverse problems},
  author={Aggarwal, Hemant K and Mani, Merry P and Jacob, Mathews},
  journal={IEEE transactions on medical imaging},
  volume={38},
  number={2},
  pages={394--405},
  year={2018},
  publisher={IEEE}
}

@article{wang2022metateacher,
  title={MetaTeacher: Coordinating Multi-Model Domain Adaptation for Medical Image Classification},
  author={Wang, Zhenbin and Ye, Mao and Zhu, Xiatian and Peng, Liuhan and Tian, Liang and Zhu, Yingying},
  journal={Advances in Neural Information Processing Systems},
  volume={35},
  pages={20823--20837},
  year={2022}
}

@article{zhang2023spatially,
  title={Spatially Covariant Lesion Segmentation},
  author={Zhang, Hang and Wang, Rongguang and Zhang, Jinwei and Liu, Dongdong and Li, Chao and Li, Jiahao},
  journal={arXiv preprint arXiv:2301.07895},
  year={2023}
}

@inproceedings{han2021redet,
  title={Redet: A rotation-equivariant detector for aerial object detection},
  author={Han, Jiaming and Ding, Jian and Xue, Nan and Xia, Gui-Song},
  booktitle={Proceedings of the IEEE/CVF Conference on Computer Vision and Pattern Recognition},
  pages={2786--2795},
  year={2021}
}

@inproceedings{teed2020raft,
  title={Raft: Recurrent all-pairs field transforms for optical flow},
  author={Teed, Zachary and Deng, Jia},
  booktitle={Computer Vision--ECCV 2020: 16th European Conference, Glasgow, UK, August 23--28, 2020, Proceedings, Part II 16},
  pages={402--419},
  year={2020},
  organization={Springer}
}

@article{kong2021breaking,
  title={Breaking the dilemma of medical image-to-image translation},
  author={Kong, Lingke and Lian, Chenyu and Huang, Detian and Hu, Yanle and Zhou, Qichao and others},
  journal={Advances in Neural Information Processing Systems},
  volume={34},
  pages={1964--1978},
  year={2021}
}

@article{zhang2023deda,
  title={DeDA: Deep Directed Accumulator},
  author={Zhang, Hang and Wang, Rongguang and Hu, Renjiu and Zhang, Jinwei and Li, Jiahao},
  journal={arXiv preprint arXiv:2303.08434},
  year={2023}
}

@inproceedings{paris2006fast,
  title={A fast approximation of the bilateral filter using a signal processing approach},
  author={Paris, Sylvain and Durand, Fr{\'e}do},
  booktitle={Computer Vision--ECCV 2006: 9th European Conference on Computer Vision, Graz, Austria, May 7-13, 2006, Proceedings, Part IV 9},
  pages={568--580},
  year={2006},
  organization={Springer}
}

@article{kirillov2023segment,
  title={Segment anything},
  author={Kirillov, Alexander and Mintun, Eric and Ravi, Nikhila and Mao, Hanzi and Rolland, Chloe and Gustafson, Laura and Xiao, Tete and Whitehead, Spencer and Berg, Alexander C and Lo, Wan-Yen and others},
  journal={arXiv preprint arXiv:2304.02643},
  year={2023}
}

@article{ma2023segment,
  title={Segment anything in medical images},
  author={Ma, Jun and Wang, Bo},
  journal={arXiv preprint arXiv:2304.12306},
  year={2023}
}

@inproceedings{zhang2017bilinear,
  title={Bilinear lithography hotspot detection},
  author={Zhang, Hang and Zhu, Fengyuan and Li, Haocheng and Young, Evangeline FY and Yu, Bei},
  booktitle={Proceedings of the 2017 ACM on International Symposium on Physical Design},
  pages={7--14},
  year={2017}
}

@inproceedings{zhang2016enabling,
  title={Enabling online learning in lithography hotspot detection with information-theoretic feature optimization},
  author={Zhang, Hang and Yu, Bei and Young, Evangeline FY},
  booktitle={2016 IEEE/ACM International Conference on Computer-Aided Design (ICCAD)},
  pages={1--8},
  year={2016},
  organization={IEEE}
}

@inproceedings{ronneberger2015u,
  title={U-net: Convolutional networks for biomedical image segmentation},
  author={Ronneberger, Olaf and Fischer, Philipp and Brox, Thomas},
  booktitle={Medical Image Computing and Computer-Assisted Intervention--MICCAI 2015: 18th International Conference, Munich, Germany, October 5-9, 2015, Proceedings, Part III 18},
  pages={234--241},
  year={2015},
  organization={Springer}
}

@article{chan2001active,
  title={Active contours without edges},
  author={Chan, Tony F and Vese, Luminita A},
  journal={IEEE Transactions on image processing},
  volume={10},
  number={2},
  pages={266--277},
  year={2001},
  publisher={IEEE}
}

@article{nie2021deformable,
  title={Deformable image registration based on functions of bounded generalized deformation},
  author={Nie, Ziwei and Li, Chen and Liu, Hairong and Yang, Xiaoping},
  journal={International Journal of Computer Vision},
  volume={129},
  pages={1341--1358},
  year={2021},
  publisher={Springer}
}

@article{chen2022joint,
  title={Joint segmentation and discontinuity-preserving deformable registration: Application to cardiac cine-MR images},
  author={Chen, Xiang and Xia, Yan and Ravikumar, Nishant and Frangi, Alejandro F},
  journal={arXiv preprint arXiv:2211.13828},
  year={2022}
}

@article{islam2020much,
  title={How much position information do convolutional neural networks encode?},
  author={Islam, Md Amirul and Jia, Sen and Bruce, Neil DB},
  journal={arXiv preprint arXiv:2001.08248},
  year={2020}
}

@inproceedings{zhang2019multiple,
  title={Multiple sclerosis lesion segmentation with tiramisu and 2.5 d stacked slices},
  author={Zhang, Huahong and Valcarcel, Alessandra M and Bakshi, Rohit and Chu, Renxin and Bagnato, Francesca and Shinohara, Russell T and Hett, Kilian and Oguz, Ipek},
  booktitle={Medical Image Computing and Computer Assisted Intervention--MICCAI 2019: 22nd International Conference, Shenzhen, China, October 13--17, 2019, Proceedings, Part III 22},
  pages={338--346},
  year={2019},
  organization={Springer}
}

@article{zhang2018road,
  title={Road extraction by deep residual u-net},
  author={Zhang, Zhengxin and Liu, Qingjie and Wang, Yunhong},
  journal={IEEE Geoscience and Remote Sensing Letters},
  volume={15},
  number={5},
  pages={749--753},
  year={2018},
  publisher={IEEE}
}

@article{he2022fully,
  title={Fully transformer network for skin lesion analysis},
  author={He, Xinzi and Tan, Ee-Leng and Bi, Hanwen and Zhang, Xuzhe and Zhao, Shijie and Lei, Baiying},
  journal={Medical Image Analysis},
  volume={77},
  pages={102357},
  year={2022},
  publisher={Elsevier}
}

@article{codella2019skin,
  title={Skin lesion analysis toward melanoma detection 2018: A challenge hosted by the international skin imaging collaboration (isic)},
  author={Codella, Noel and Rotemberg, Veronica and Tschandl, Philipp and Celebi, M Emre and Dusza, Stephen and Gutman, David and Helba, Brian and Kalloo, Aadi and Liopyris, Konstantinos and Marchetti, Michael and others},
  journal={arXiv preprint arXiv:1902.03368},
  year={2019}
}

@article{vercauteren2009diffeomorphic,
  title={Diffeomorphic demons: Efficient non-parametric image registration},
  author={Vercauteren, Tom and Pennec, Xavier and Perchant, Aymeric and Ayache, Nicholas},
  journal={NeuroImage},
  volume={45},
  number={1},
  pages={S61--S72},
  year={2009},
  publisher={Elsevier}
}

\end{document}